\documentclass{article}
\usepackage{graphicx} 

\usepackage[utf8]{inputenc}
\usepackage[T1]{fontenc}
\usepackage[english]{babel}
\usepackage{amsmath,amssymb}
\usepackage{booktabs}
\usepackage{multirow}
\usepackage{array}
\usepackage{tabularx}
\usepackage[section]{placeins}  
\usepackage{xcolor}
\usepackage{pifont}            
\usepackage[ruled,linesnumbered,noend]{algorithm2e}
\usepackage{listings}
\usepackage{url}
\usepackage{cite}
\usepackage[margin=2cm]{geometry}
\usepackage[colorlinks=true,linkcolor=blue,citecolor=blue,urlcolor=blue]{hyperref}

\title{EvoScore: LLM-Guided Evolutionary Program Search for Online Two-Dimensional Strip-Packing Heuristics}
\author{Mariusz Kaleta, Waldemar Kołodziejczyk, Izabela Zoltowska}
\date{October 2026}

\begin{document}

\maketitle

\section{Introduction}\label{sec:introduction}
The Two-Dimensional Online Rectangular Strip Packing Problem (2D-OSPP) is a classical combinatorial optimization problem widely studied in the literature. It aims to place axis-aligned rectangular items into a strip of fixed width and unbounded height to minimize the total height, or equivalently, to maximize strip utilization. Items arrive sequentially, and each item must be placed immediately upon arrival without overtaking previously placed items. The 2D-OSPP forms the foundation of numerous industrial applications, particularly in automated storage and retrieval systems; manufacturing and cutting-stock operations involving paper, steel, glass, wood, and textiles (where the strip represents a stock roll or sheet); mixed-pallet and container loading; trim-loss minimization; and warehouse slotting of rectangular footprints \cite{ref1, ref2, ref3}. Since practical cases routinely operate on massive volumes, even marginal improvements in strip utilization yield substantial cost savings in raw materials, freight, and warehousing.

The online aspect involves making fast decisions under limited knowledge of rectangles arriving in the future. Therefore, practical approaches primarily rely on fast online heuristics. Classical online heuristics enumerate a finite set of candidate corner placements to find the best position for the arriving rectangle \cite{ref4}. They select the best placement using a scalar scoring rule. The score is calculated separately for each candidate placement, and a heuristic selects the corner with the best score. For instance, the Bottom-Left heuristic chooses the lowest, then leftmost, corner \cite{ref5}; An online adaptation of the Best-Fit idea of Burke et al. prefers the placement that increases the current strip height least \cite{ref6}; the MaxRects and skyline families rank corners by criteria such as the lowest resulting top edge or the best short-side fit \cite{ref7}; and contact-based rules, such as the Touching Perimeter criterion introduced by Lodi et al. \cite{ref8}, maximize the perimeter shared with neighbors. According to the No Free Lunch theorem, no single heuristic dominates across all distributions of rectangle sizes and shapes \cite{ref9}. Therefore, in a given context, one tries to choose or design the best-suited rule tuned to a specific item distribution. Such a process of tailoring a heuristic is typically slow, manual, and time-consuming, and is rarely done; as a result, off-the-shelf heuristics are often chosen intuitively. 


In this paper, we propose a method to automate heuristic tailoring by representing the selection rule as a short function in a programming language that calculates real-valued scoring for a given candidate placement. Such a scoring function undergoes evolutionary improvement within a loop, utilizing a large language model (LLM) to generate proposed changes to the function. This approach falls within an area of active research focused on automated algorithm discovery.

Our work originates mainly from two previous achievements: 1) our previous work on solving the online 2D strip packing problem with Reinforcement Learning and a transformer-based Agent \cite{ref11}, and 2) the FunSearch framework searching through the space of computer code through an evolutionary process controlled by an LLM \cite{ref10}. In \cite{ref11}, we introduced a framework in which a placement of a newly arrived item is selected with a learned policy from a finite set of candidates resulting from a so-called MaxRects representation. That approach allowed for discovering better heuristics adapted to specific characteristics of problem classes, but was represented with a neural network that is hard to interpret. On the other hand, FunSearch has demonstrated that it is possible to discover better heuristics through an LLM-driven program-evolution process for one-dimensional bin-packing problems \cite{ref10}, and heuristics represented with computer program code retain a high degree of potential interpretability. However, the approach achieved for one-dimensional bin-packing problems does not transfer directly to two dimensions, as scalar scoring becomes insufficient to reflect relative differences between candidates in the more complex geometry of 2D positioning and element orientations, which together constitute a continuous space of candidates.

To achieve lower resource consumption and greater interpretability of the discovered heuristic than previous transformer-based policy approaches, we follow and further develop the FunSearch concept. The resulting EvoScore improves upon previous packing applications that utilize a continuous candidate space by first generating a fixed set of candidates, then allowing the scoring function to access all candidates when scoring a single candidate. This approach enables normalization, relative ranking, and comparative reasoning prior to selection. Moreover, our EvoScore departs from FunSearch in the following four essential ways: 1) \textit{Classical warm starts}, which means seeding the islands of the search from five different classical heuristics rather than from one program; 2) \textit{Model switching}, involving dynamically switching the active LLM in case of stagnation or malformed output contrary to using a single static model; 3) \textit{Context-conditioned prompting} that includes information on the instance distribution in each prompt; and 4) \textit{Validation floor}, that is, a selection of the heuristic based on three disjoint sets of instances, and acceptance of the evolved program only if it outperforms the best classical baseline solution on a set of instances not used during the learning process. Instead of aiming at one universal heuristic, EvoScore targets the discovery of heuristics suited to specific instance classes. We consider 72 instance classes depending on item size distributions and episode lengths.

Our key contributions are threefold:
\begin{itemize}
\item	We demonstrate that it is possible to automatically discover heuristics through LLM-guided evolution for 2D-OSPP that are tailored for instance classes. For that, we propose a specific representation of the search space in the form of a scoring function over a canonical set of candidate corner placements. 
\item	We propose several new improvement techniques to the standard FunSearch approach, such as \textit{Classical warm starts}, \textit{Model switching}, \textit{Context-conditioned prompting}, and \textit{Validation floor}. These techniques enabled us to achieve significantly better results than each of the eleven online baselines we tested in paired tests over 7,200 episodes,  winning 60 of the 72 instance classes against the best competitor in each class. Moreover, analysis reveals a consistent pattern: evolution keeps the seed's primary objective and adds one to three interpretable tie-break terms.
\item We provide empirical evidence that the distribution of parameters is sufficient. We demonstrate that evolution trained purely on fitted distribution parameters performs statistically equivalent (within a one-percentage-point margin) to direct evolution on real logistics data, showing that heuristics can be synthesized from summary statistics alone.
\end{itemize}

The rest of the paper is organized as follows. Section \ref{sec:relatedWork} reviews related work and highlights existing research gaps.  We conclude this section with a summary of the
contribution of this paper. Section \ref{sec:method} outlines our methodology, mainly the EvoScore approach. Section \ref{sec:design} describes our experimental design. Section \ref{sec:results} presents results on the synthetic benchmark and real data, followed by a discussion in Section \ref{sec:discussion}. Finally, Section \ref{sec:conclusions} concludes the paper and displays limitations and future work.
\section{Related work}\label{sec:relatedWork}
\subsection{Online two-dimensional strip packing and classical heuristics}

Optimization of packing problems has a long history, and the online version has its own challenges; see, e.g., the recent review by \cite{hifi2026critical}. 
Algorithms that make packing decisions aim to minimize the strip's height, or equivalently, maximize the strip's area utilization. The online, irrevocable regime arises from several real-world applications, such as material-handling and cutting processes.  The problem has been extensively studied, resulting in numerous algorithms, starting with the shelf algorithms by \cite{ref14}, further enhanced and revisited by \cite{ref15} and \cite{ref17}.  In a more recent review \cite{christensen2017approximation} focused on online algorithms for bin packing, the results on the performance of several important algorithms are briefly summarized, including the 2D-OSPP case, where a high gap between the best upper and lower bounds remains an open problem.

On the other hand, \cite{ref3} provides a recent overview of methods for handling practical constraints in the strip-packing problem, with unloading order among the most common requirements. 
The connection between unloading constraints and online packing is highlighted by \cite{wei_open_2019}: the packing sequence in both cases dictates the feasibility of the spatial layout,
but in the online setting it is unknown (or only partially known in semi-online approaches \cite{Boyar2017}). 
Although there are a limited number of studies that have employed exact methods to solve formal mathematical programming models that integrate unloading constraints, most studies apply heuristic algorithms, as surveyed in \cite{oliveira2016survey}. The constructive heuristics are most common. These include 
fitness-based heuristics such as Best-Fit, and positioning-based
such as Bottom-Left, as classified by \cite{oliveira2016survey}.

A standard practice applied together with heuristic placement rules is to restrict the potential placements to corner points
\cite{ref54},  which Crainic et al. \cite{ref4} generalize to extreme points. 
Similarly, MaxRects introduced by Jylänki (2010) \cite{ref7}, represents space available for packing as a set of
free rectangles whose corners implicitly include extreme points.
Restricting the placement possibilities to MaxRects does not exclude any solutions achievable by typical baseline heuristics, while reducing the continuous decision space to a finite set of candidates. Several metaheuristic methods take advantage of MaxRects representation of possible placements, e.g., \cite{wong2024integrating, xiong}, and our earlier work \cite{ref11}.

For the three-dimensional online case, \cite{ref21} reviews offline and online solution approaches and proposes a family of online heuristics, along with an algorithm-selection framework, for the semi-online setting with full lookahead. Again, classical heuristics that can be simply transformed into online rules, such as Bottom-Left \cite{ref5}, Best-Fit \cite{ref6}, and the skyline and MaxRects families \cite{ref7}, remain the practical default because they are fast and interpretable.

\subsection{Automated heuristic design and algorithm selection}

Automating the design of decision rules, rather than handcrafting specific heuristics, is a promising trend in combinatorial optimization, including cutting and packing problems \cite{Mezghani2023}. Previously, researchers demonstrated the capabilities of genetic algorithms
to search the space of
heuristics and derive automated frameworks. One example in online bin packing is the CHAMP framework (Creating Heuristics via Many Parameters) \cite{asta2016champ}, deciding which bin to choose for the current item based on a heuristic score specialized with a genetic algorithm to the given
distributions.
Another example is using genetic algorithm evolution to develop hyper-heuristics that provide dispatching rules for production scheduling; this line is consolidated in the survey by \cite{ref22}. To the best of our knowledge, no automated heuristic design has been proposed for packing problems that require specifying detailed placement decisions, i.e., in terms of items' coordinates.

In this regard, a complementary line of research falls into a category of algorithm selection problems. Previously mentioned work by \cite{ref21}
developed a pool of heuristics and used supervised machine learning techniques to predict the most promising heuristic for solving any new instance from a specific class of problems. In this way, heuristics are not automatically created, but are automatically chosen from the larger pool of candidates.  Our earlier work \cite{ref11} followed a similar path but used a transformer policy to learn to choose the best heuristic for scoring placement decisions.

All the aforementioned approaches draw from 
No Free Lunch theorems \cite{ref9}, 
clearly highlighting that no single method is best for all instances and the most effective policies come from matching the method to the problem class.
To further develop problem-specific algorithms that require no human intervention, new approaches aim to leverage Large Language Models (LLMs) as code generators and variation operators in program search.

\subsection{LLM-driven heuristic design}
\label{LLM_driven_heuristics}

A rapidly expanding line of research uses LLMs to automate heuristic design \cite{liu2026systematic}. One of the first well-known examples is the FunSearch framework \cite{ref10}, which pairs a pretrained LLM with an evolutionary search loop over an island-structured database of heuristics stated as priority rules for decisions in combinatorial problems.
It shows that LLMs can discover new constructive heuristics that outperform classical baselines such as First-Fit and Best-Fit on problems such as the cap set problem and 1D online bin packing.

Subsequent frameworks refined the evolutionary loop introduced in FunSearch. EoH \cite{ref25} 
makes an attempt toward the previously mentioned 
 automated heuristic design: 
it uses both natural-language ``thoughts'' and code to represent heuristics. 
 ReEvo \cite{ref26} introduces short- and long-term reflective feedback to guide the search direction; and EvoTune \cite{ref27} 
 demonstrates that  continuously refining 
 the LLM policy  through reinforcement learning 
 accelerates the discovery of superior algorithms.
 
 More recent agentic and multi-LLM variants incorporate role-based collaboration among prompted agents \cite{ref28}, reinforcement learning governing the design strategy itself \cite{ref29}, and state-dependent heuristic selection combined with code evolution \cite{ref30}. Beyond packing, LLM-evolved dispatching rules have demonstrated success in dynamic shop scheduling \cite{ref31, ref32} and single-machine sequencing \cite{ref33}.

Despite these advances, empirical evaluations highlight critical limitations that shape our method's design. First of all, \cite{ref35} critically analyzed online bin packing heuristics developed using LLM-based methods, highlighting that they are hardly interpretable and unable to generate novel algorithmic concepts. Moreover,
\cite{ref34} benchmarked LLM-discovered bin-packing heuristics across broad instance sets and showed poor generalization of evolved rules when applied beyond their training distributions. Furthermore, \cite{ref37} demonstrated that simple hand-crafted heuristics can match LLM-generated rules, but are far more understandable. 
These limitations add to the observation in a survey paper by \cite{liu2026systematic} that the tasks considered by LLM-guided search processes are relatively simple: in the field of packing problems, the discoveries were limited mainly to 1D bin selection problems, so specifying detailed placement decisions was not required.

Consideration of multidimensional packing with LLM-guided program search remains nascent; for example, only three papers are identified in a survey by \cite{hifi2026critical}. 
First,
\cite{ref36} 
demonstrate the viability of prompted LLM evolution for 2D bin packing,
noting, however, that their approach lacks systematic accounting for LLM non-determinism across multiple runs and sensitivity to prompt engineering variations. On the other hand, previously mentioned work by 
\cite{ref37}
investigated the ability of LLMs to generate a complete solver for the constrained 3D packing problem.
They observed that LLMs almost exclusively focus their evolutionary search on refining the candidate scoring function, yielding a heuristic whose performance matches that of human-designed greedy algorithms. 
In their work,  
scoring functions are part of the overall algorithm invented and refined by the LLM; they evaluate each valid candidate position separately. 
In the end,
their framework relies heavily on hand-crafted code to prevent execution failure and struggles under highly constrained problem settings.
Finally, \cite{yang2025learning}
investigated the online 3D robotic packing system and proposed a
packing-and-unpacking synergistic network that
uses LLM-generated heuristics to guide deep reinforcement learning (DRL) explorations. The space states are defined over a grid representing the bin, so the problem size is highly limited.  
Moreover, few details are provided about the LLM, making it difficult to replicate.
 To our knowledge, LLM-guided program search has not been previously explored for online 2D strip packing.

 Finally, to establish the provenance of our architectural choices in Section \ref{sec:search}, we examine several relevant strands of prior approaches on prompted LLM evolution below.
 \begin{itemize}
     \item {\bf Initial code seeding.} 
     FunSearch \cite{ref10} seeds 
     one initial program in every island
     considered.
     For the bin-packing task, this initial program implements the Best-Fit heuristic. Seeding the search from an existing heuristic has since become common practice: EvoTune also starts from the Best-Fit \cite{ref27}, Karimi et al. seed every island with a base heuristic \cite{ref44}, ReEvo accepts trivial or expert seed functions \cite{ref26}. and Huang et al. \cite{ref41} show that the code example 
     Valuable insight from subsequent studies, i.e., \cite{ref41},demonstrates that seed code example provided in the prompt influences the generated program the most.   
      Highly motivating \cite{ref42} propose the Hercules algorithm, which abstracts the core components from elite heuristics to incorporate them as prior knowledge in prompts.
      \item {\bf Generator model ensembles.}
      Most existing frameworks rely on a single generator model or assign distinct roles to instances of the same model \cite{ref28}. 
      Since FunSearch publication, the use of multi-model pools has emerged, such as AlphaEvolve \cite{ref45} employing a fixed pairing of a fast and a strong model, or ShinkaEvolve \cite{ref46} utilizing a bandit algorithm 
       to dynamically select models based on fitness improvement, allowing to surpass both uniform ensembles and single-model baselines.
       To our knowledge, no prior work dynamically switches active generator models in response to syntax errors or search stagnation.
      \item {\bf Distribution-conditioned generation.}
Standard prompts, as desinged in FunSearch, EoH and ReEvo carry only code skeletons and elite exemplars,
      failing to account for the heterogeneity across instances. 
      Here, we follow the concept of instance-specific heuristic generation proposed by 
      \cite{ref47}, which divides problem into subclasses using instance features, and provides a description of each sub-class in the prompt.
      \item {\bf Held-out validation.} While FunSearch \cite{ref10} evaluates and selects the generated heuristic that achieves the best results on training instances before reporting out-of-sample results, subsequent frameworks explicitly target generalization failures. 
      The recently introduced EoH-S \cite{ref43} framework evolves complementary sets of heuristics to optimize diverse problem instances, whereas \cite{ref44} exposes the generator directly to instances on which existing heuristics have underperformed.
      Selecting candidate heuristics using a dedicated validation set has also proven effective in discovering vehicle routing operators \cite{ref48}. Here, we follow the same approach. Additionally, not common in LLM-driven online heuristic design, we
      employ an explicit fallback mechanism to the incumbent baseline, which is inspired by work \cite{ref49} devoted to constraint-model reformulation, where 
      returning to the original model if no candidate outperforms it has been shown to be an efficient strategy.
      \item {\bf Execution environments.} Sandboxed code execution, first established in FunSearch \cite{ref10}, has been enhanced in later systems by staged evaluation \cite{ref45} and population-count-based island resets \cite{ref50}. The underlying FunSearch paradigm continues to drive applications in deletion-correcting codes \cite{ref50}, Feynman integral reduction \cite{ref51}, unit commitment \cite{ref52}, and traffic engineering \cite{ref44}, while also providing the basis for general-purpose search platforms such as LLM4AD \cite{ref53}.
   \end{itemize}

 \subsection{Neural and reinforcement-learning approaches to strip packing}\label{sub_neural}

 A parallel line of research uses artificial intelligence for strip packing without using LLMs. There are several approaches that use neural networks; for example, \cite{ref38} trained compact neural networks via black-box optimization and reported up to +6.74 percentage points over the best MaxRects and skyline variants, which was further improved in \cite{ref39} via per-instance local search in the space of network weights. Similarly, GraphPack \cite{ref40} applies a black-box reinforcement learning policy to learn item selection and candidate placement simultaneously. However, these methods address the offline problem, in which the algorithm also chooses the packing sequence.

   Our prior work \cite{ref11} addressed the online setting of strip packing problem by introducing a candidate-placement framework that selects placements from a MaxRects free-space representation using a transformer trained via proximal policy optimization. 
   A similar methodology was also used in \cite{xiong}, where empty maximal spaces provided inputs for the transformer-based DRL 
   determining the position and orientation of items in online
3D bin packing. Another work, addressing
online
3D bin packing is \cite{wong2024integrating}, where  the extreme
point heuristic was integrated into a DRL framework.
   Even though the reported machine-learning attempts proved their high performance and generalization capabilities, they require GPU training per distribution, rely on neural inference at runtime, and yield decision rules that are hard to explain.

   \subsection{Summary and research gaps}

   The literature review identifies three primary research gaps guiding
   the paper's main contributions:

   \begin{description}
       \item[LLM-driven design of heuristics for the 2D-OSPP:] While LLM-driven program search succeeds in 1D bin packing \cite{ref10, ref25, ref27, ref43} and dynamic scheduling \cite{ref31, ref32}, its ability to
       design heuristics for the online 2D strip packing remains unproven. 
       We address that complicated problem by leveraging the widely observed ability of LLM-guided search processes to refine scoring functions \cite{liu2026systematic}. We propose a novel design for LLM-driven heuristic search that mitigates underlying  complexity through an appropriate, human-designed way of using and representing the scoring functions. 
       \item[Methodological gaps 
       in LLM-driven heuristic generation: ]
       Despite rapid interest in LLM-driven heuristic design, current methods remain in an early stage of maturity and rely on naive evolutionary setups. For instance,
       LLM-evolved heuristics 
       could be improved by incorporating basic machine learning practices, such as held-out validation floors, dynamic fallback mechanisms to baseline incumbents, or heterogeneous model-switching strategies during search stagnation.
       Moreover, the  
       code attribution relative to initial seeds \cite{ref35} is not well recognized. Consequently, 
       well-established optimization strategies such as classical warm starts are not thoroughly recognized in the LLM-driven heuristic design.
       Current methodologies remain vulnerable to out-of-sample performance degradation, and under-equipped to consistently outperform strong classical baselines.
       Our contribution addresses this gap by proposing several improvement techniques, while providing a program-level attribution analysis. 
       \item[Search data requirements of particular instances versus fitted distributions.] 
        Prior frameworks often optimize directly on target instances without examining whether evolutionary gains stem from specific instance structures or summary distribution statistics \cite{ref10,ref25, ref37}. Our contribution lies in systematically comparing search performance on direct instances versus parametric fitted distributions.
   \end{description}
\section{EvoScore}
\label{sec:method}

\subsection{Problem statement}
\label{sec:problem}

We consider a strip of width $W$ and a sequence of $n$ rectangular items with widths and heights $(w_i, h_i)$, $i = 1, \dots, n$. Items arrive one at a time, and the following are not visible to the packing agent. When item $i$ arrives, its dimensions become known, and a placement must be chosen immediately, using only the information available at that moment, i.e., the dimensions and placements of items $1, \dots, i-1$ and the dimensions of item $i$; the placement cannot be changed afterward. A placement is a triple $(x_i, y_i, r_i)$, where $x_i$ and $y_i$ represent the x and y coordinates of the bottom-left corner of item $i$, and $r_i \in \{0, 1\}$ indicates a rotation of 90 degrees, so the oriented dimensions are $(w_i', h_i') = (w_i, h_i)$ if $r_i = 0$ and $(h_i, w_i)$ otherwise. The placement must fit within the strip, resulting in the following constraints:
\begin{equation}
x_i \ge 0, \qquad y_i \ge 0  \qquad x_i + w_i' \le W,
\label{eq:inside}
\end{equation}
and must not overlap any previously placed item $j < i$, i.e., the open rectangles $(x_i, x_i + w_i') \times (y_i, y_i + h_i')$ and $(x_j, x_j + w_j') \times (y_j, y_j + h_j')$ are disjoint. After all items have been placed, the strip height $H$ equals:
\begin{equation}
H = \max_{i} \,(y_i + h_i'),
\label{eq:height}
\end{equation}
and the objective is to minimize $H$, or, equivalently, to maximize the strip utilization.  
The search in our framework aims to maximize utilization, computed as:
\begin{equation}
U = \frac{\sum_{i=1}^{n} w_i h_i}{W \cdot H} \in (0, 1].
\label{eq:util}
\end{equation}

Additionally, item lengths are expressed throughout the paper as fractions of the strip width, i.e., $W = 1$.

\subsection{Candidate placements and classical heuristics as selection rules}
\label{sec:candidates}
Our framework applies heuristics to select the best placement for the current item from a continuously updated set of candidate placements. In that way, both the evolved heuristics and their baseline seeds operate on identical principles.   
Specifically, the available, remaining space of the already filled strip is maintained as a set of maximal free rectangles (MaxRects) representation \cite{ref7}. When an item arrives, each free rectangle that the item fits in, in either orientation, yields four candidate placements -- one per each corner of the rectangle. Candidate placements can be numerous, so they are filtered to retain only those that ensure contact with an already placed item or a strip wall. Duplicates are removed, and the list is sorted bottom-up and left-to-right by $(y, x, w, h)$, and it is capped at the first $C$ entries in that order (the value of $C$ is given in Section~\ref{sec:benchmark}). 

The baseline heuristics, considered here as seeds for the LLM search, use different selection rules for the best placement among candidates, as summarized in Table~\ref{tab:seeds}. Namely, these rules translate into evaluating placements by assigning scores to each of them.
The first four are the familiar textbook rules. The fifth, Relative-Rank, applies the same criterion as Lowest-Top, i.e., the bottom-left rule of the MaxRects family \cite{ref7}, but in ordinal form: each candidate is scored by the rank of its top edge among all candidates, with the rank of its $x$-coordinate as the tie-break. This form implements the lexicographic order exactly and is independent of the scale of the item dimensions, whereas the normalized form of Lowest-Top is a weighted sum. We include both because they demonstrate the two elementary uses of the candidate set, normalization and ranking, that the representation of Section~\ref{sec:representation} makes available.
The difference between the scoring rules applied by the heuristics is especially important when analyzing which arguments they consider. For instance, Bottom-Left, Best-Fit and Max-Contact compute their score from the candidate and the placed items alone; each candidate is judged in isolation. Lowest-Top and Relative-Rank, by contrast, compute quantities that only make sense relative to the other candidates, i.e., they determine a normalized position within the range of top edges, or provide a rank. A scoring function that receives only the specific candidate under consideration at a time cannot express the latter, which is why we design the signature of each seed function to be evolved, so as to receive also the whole candidate set.

\subsection{The scoring function representation}
\label{sec:representation}

Each baseline heuristic serves as a seed for the evolved scoring function, and they all share the same signature. Its inputs include the strip width, the collection of previously placed items (defined by their positions and oriented sizes), the specific candidate being evaluated (its position and oriented size), and the exhaustive list of all possible candidates for the current item. 
The function outputs a real number, with higher scores indicating better placements. The runner evaluates the function on every candidate and places the item at the arg-max position.

Exemplary representations of two heuristics,  Best-Fit and Lowest-Top, are provided as follows. Best-Fit is implemented as:
\begin{lstlisting}
def score_placement(strip_width, placed_items, candidate, all_candidates):
    current_top = max((p["y"] + p["h"] for p in placed_items), default=0)
    height_increase = max(0, candidate["y"] + candidate["h"] - current_top)
    return -(height_increase * 10000 + candidate["y"] * 100 + candidate["x"])
\end{lstlisting}
and Lowest-Top as:
\begin{lstlisting}
def score_placement(strip_width, placed_items, candidate, all_candidates):
    top = candidate["y"] + candidate["h"]
    min_top = min(c["y"] + c["h"] for c in all_candidates)
    max_top = max(c["y"] + c["h"] for c in all_candidates)
    span = max_top - min_top + 1
    return -(top - min_top) / span * 10000 - candidate["x"]
\end{lstlisting}

The large constant weights enforce a lexicographic order: the primary criterion dominates, while the remaining terms break ties. Similarly, the Bottom-Left seed is a single line of the same form, Relative-Rank has seven lines, and Max-Contact, which sums shared edge lengths over the placed items, has about twenty lines. 
Besides using these heuristics as seeds, we also use them 
as a benchmark baseline (Section~\ref{sec:benchmark}).

The inclusion of the complete list of candidate placements is the one deliberate departure from a typical, per-candidate priority function. This design allows heuristics to normalize, rank, or evaluate a candidate relative to the entire pool of alternatives. As we demonstrate in Section~\ref{sec:results-seeds}, not only do the Lowest-Top and Relative-Rank seeds capitalize on this feature, but all the evolved programs also rely heavily on it to resolve ties.

We should also note that generating the candidate set outside the evolved function offers an additional advantage beyond limiting the decision space. By decoupling the geometric generation of candidates from the decision rule used to score them, the evolved function is spared from reasoning over raw spatial coordinates. Instead, it only needs to rank a limited set of viable options.  Since each seed is itself a program of this form, the seeds serve both as the initial population of the search and as the first few-shot examples, which show the LLM exactly the kind of object it is asked to improve.

The final note should point to the code representation of heuristics. Although simple, it could still be stated in several semantically equivalent and syntactically different ways. Since the LLM edits text, the chosen representation influences the search starting point. We therefore wrote each seed in the most direct form possible, using one named quantity per line, so that the few-shot examples clearly express the implemented rule. Whether other implementations of the same rule would lead the search elsewhere is not examined in this paper.

\subsection{The LLM-guided evolutionary search}
\label{sec:search}

\begin{figure}
\centering
\includegraphics[width=0.9\linewidth]{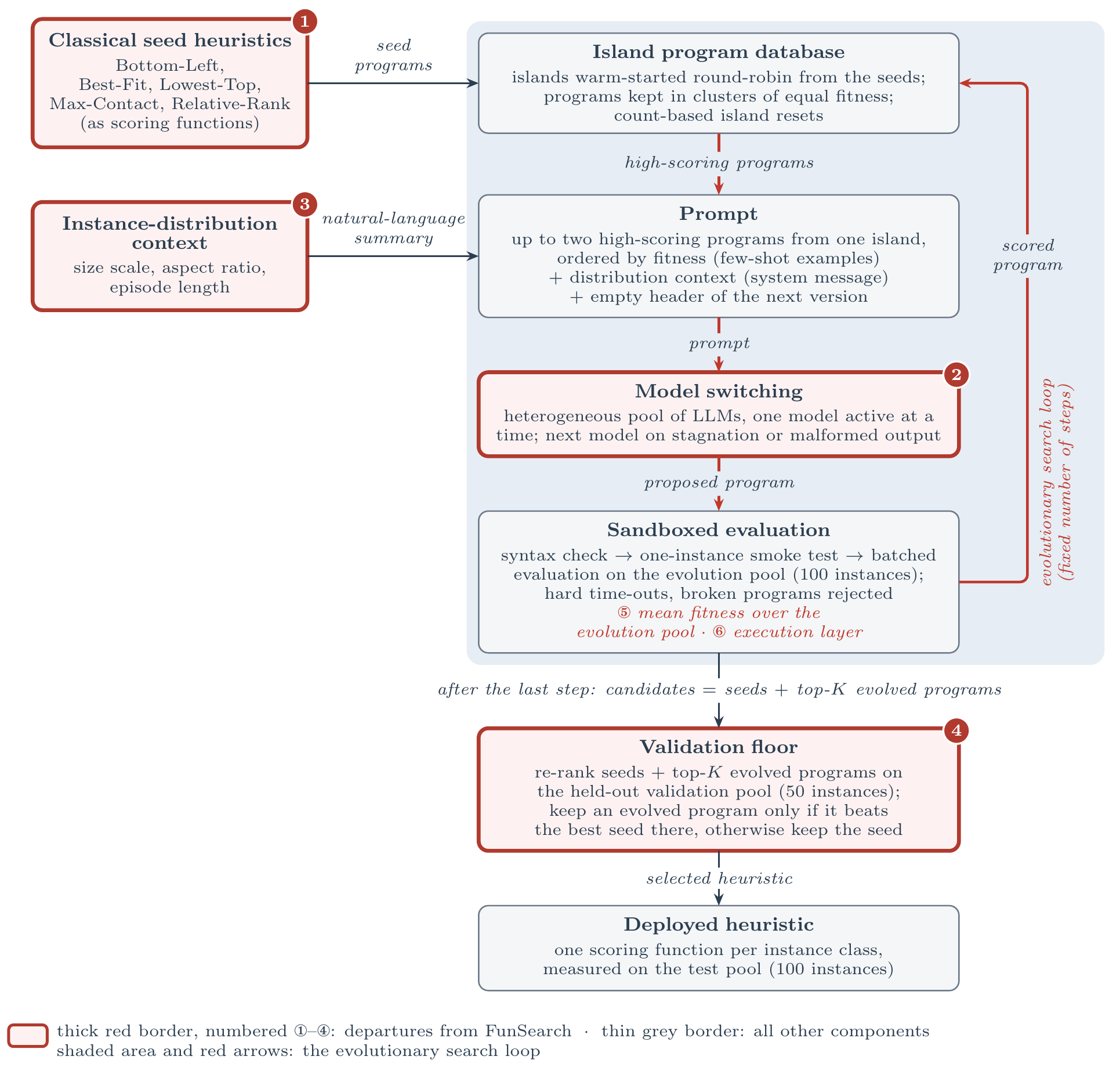}
\caption{The EvoScore pipeline. Classical warm starts \ding{172} seed an island program database from the five classical seeds; model switching \ding{173} draws edited scoring programs from a heterogeneous pool of LLMs, prompted with best examples and, through context-conditioned prompting \ding{174}, with a description of the item distribution; proposals are evaluated in a sandbox on the evolution pool and registered back in the database; after the last step, the validation floor \ding{175} decides whether an evolved program or the best seed is deployed, and all reported numbers come from a third, disjoint test pool. Thick red borders and circled numbers mark the modifications of FunSearch methodology; \ding{176} and \ding{177} are implementation details described in the text.}
\label{fig:pipeline}
\end{figure}

EvoScore inherits the core methodological design of FunSearch: an island-structured program database, sampling of best programs as few-shot context, LLM-proposed modifications of the evolved function, and an execute-score-retain loop. The components inherited unchanged are drawn with thin gray borders in Figure~\ref{fig:pipeline}, illustrating the architecture of EvoScore. The four improvements to FunSearch, applied in this work, are marked by thick red borders and the numbers \ding{172} to \ding{175}, while the fitness definition and the execution layer, marked \ding{176} and \ding{177} in the figure, are implementation details described at the end of this section.

\paragraph{\ding{172} Classical warm starts.}
To ensure initial seed variability, EvoScore distributes the five seed heuristics listed in Table~\ref{tab:seeds} across twelve islands using a round-robin approach. 
These baseline heuristics, in addition to seeding, also serve as the benchmark for newly evolved programs.
This is the basis of the analysis in Section~\ref{sec:results-seeds}, which offers some explanations of the evolved rules, taking into consideration the need for interpretability and contextualization raised by Herrmann and Pallez \cite{ref35}.

\begin{table*}[!tbp]
\centering
\caption{The five seeds, written as selection rules over the candidate set.}
\label{tab:seeds}
\small
\begin{tabularx}{\textwidth}{>{\raggedright\arraybackslash}p{4.3cm}>{\raggedright\arraybackslash}X>{\raggedright\arraybackslash}p{3.6cm}>{\raggedright\arraybackslash}p{3.0cm}}
\toprule
Heuristic & Primary criterion (best first) & Tie-break & Looks at other candidates? \\
\midrule
Bottom-Left \cite{ref5} & lowest $y$ & lowest $x$ & no \\
Best-Fit \cite{ref6} & smallest increase of the current strip height & lowest $y$, then lowest $x$ & no (uses placed items) \\
Lowest-Top (MaxRects-BL criterion, normalized form) \cite{ref7} & lowest top edge $y + h$, normalized over the candidate set & lowest $x$ & yes (normalization) \\
Max-Contact \cite{ref7,ref8} & longest perimeter shared with the strip walls and the edges of placed items & none & no (uses placed items) \\
Relative-Rank  (MaxRects-BL criterion, rank form) \cite{xiong} & rank of top edge among candidates & rank of $x$ among candidates & yes (ranks) \\
\bottomrule
\end{tabularx}
\end{table*}

\paragraph{\ding{173} Model switching.}
To overcome program evolution problems, such as syntax errors or search stagnation, EvoScore draws from an ordered pool of three low-cost models, accessed via a hosted API: DeepSeek V4 Flash, Gemini 2.5 Flash Lite, and Qwen3 Coder. The purpose of the pool is to introduce variance into the proposals: a single model that is sampled repeatedly tends to return variations of the same program, whereas a model from another family is likely to propose a different one. The selection was dictated by cost. Since the search issues several hundred requests per instance class, only models from the low-cost tier were considered, and among them we chose three that differ as much as possible, i.e., models of three vendors and three independently developed families, two of them general-purpose and one code-specialized. The pool is ordered by increasing price per request, so that most requests go to the cheapest model and the most expensive one is reached only when the cheaper ones have stalled. The models are not equally capable: on SWE-bench Verified, the vendors report at least 73.7\% for DeepSeek V4 Flash \cite{deepseekv4flash}, 67.0--69.6\% for Qwen3 Coder \cite{qwen3coder}, and 31.6\% for Gemini 2.5 Flash Lite in a single attempt \cite{gemini25flashlite}, figures that are obtained with different agent scaffolds and indicate an order of magnitude only. Individual strength was, however, not the selection criterion; the pool is meant to be diverse and affordable.
Every run starts with the first model at a sampling temperature of 0.8. A transition to the subsequent model of the pool, returning to the first one after the last, is triggered whenever the system encounters problems indicative of stagnation and a lack of evolution. Specific rules applied are: three consecutive prompts for which no valid scoring function is obtained, each proposal being drawn up to three times, or fifteen consecutive registered programs that fail to improve the best current fitness. Each switch also raises the sampling temperature by 0.1, up to a maximum of 1.2. The pool is intended to broaden the distribution of the proposed programs, thereby maintaining exploration despite a single model being stuck, while keeping the per-proposal cost of low-cost models. How often the switch was triggered and how the requests were distributed over the three models is reported in Section~\ref{sec:results-funsearch}.
\paragraph{\ding{174} Context-conditioned prompting.}
Each generation prompt is conditioned on a short natural-language description of the statistical problem distribution, i.e., the items' size scale, their aspect-ratio profile, and the episode length.
EvoScore evolves one heuristic per item class, and the description specializes the generator to the expected item distribution.
This approach also enables executing the search using merely the summary statistics of the item population, a capability that is leveraged in Section~\ref{sec:realdata-protocol}. In Section~\ref{sec:results-real}, we show a constant stated in this context that reappears verbatim in an evolved program.

\paragraph{\ding{175} Validation floor.}

The FunSearch loop returns the program with the best fitness among all evolution episodes. In contrast, EvoScore assesses both the original seeds and the highest-performing generated programs against a separate validation pool, electing to deploy an evolved solution only when it outperforms the top seed on this held-out data. 
We designed our approach specifically to overcome 
the generalization failure, as documented by Sim, Renau and Hart \cite{ref34}.
Without held-out validation tests we would risk that
a program winning in the evolution pool may have fitted its particular episodes.
Thus, 
without additional costs 
at deployment, we propose 
to elect the winning solution only on episodes the search has never scored. Recall, that the winning solution is chosen among the evolved programs and baseline heuristics, so 
the finally implemented rule should never underperform the existing classical heuristic 
on unseen data.

\paragraph{Fitness and execution layer (\ding{176}, \ding{177}).}
Since the problem is online, a program runs on an episode by placing its items one at a time.
The whole episode is assessed by the strip height achieved after placing the last item. Fitness of the evolved heuristic is calculated as the average of this resulting height across the entire evolution pool, with the value negated to align with the search's maximization objective - this averaging mirrors the aggregation method for evaluation inputs demonstrated in the FunSearch paper. It is evaluated in a single sandboxed call, so that pools of a hundred episodes remain tractable. Notably, prior to evaluation, each new program undergoes an additional multi-step assessment procedure. First, the raw model output undergoes formatting normalization to resolve indentation, reasoning tags, and leftover markdown syntax. Second, the normalized code enters a three-stage evaluation cascade. This process begins with a syntax check that rejects flawed formulations: empty functions or those with multiple top-level return statements; it then runs a single-episode smoke test within a crash-proof subprocess enforcing a strict time limit; and finally, it executes the comprehensive batched evaluation. Any invalid programs caught during these checks are explicitly discarded. The reference FunSearch implementation resets its islands on a wall-clock timer, which makes a run depend on the latency of the language-model service; following Weindel and Heckel \cite{ref50}, who reset after a fixed number of registered programs, EvoScore ties both the island resets and the temperature schedule of elite sampling to the step count, and every run is checkpointed so that it can be resumed.

\paragraph{Three disjoint instance pools.}
For each item class, we generate disjoint episode pools based on the class's parameter distribution. Specifically, we create three pools using different random seeds, detailed below by size and purpose:
\begin{itemize}
  \item an \textbf{evolution} pool of 100 episodes, used for the per-program fitness;
  \item a \textbf{validation} pool of 50 episodes, used only in the \emph{Validation floor} \ding{175} to select the deployed program;
  \item a \textbf{test} pool of 100 episodes, used for all the results reported in the paper.
\end{itemize}

A detailed process of running EvoScore on a specific item class is presented in Algorithm~\ref{alg:evoscore}.

\begin{algorithm}[!htbp]
\caption{EvoScore search for one instance class}
\label{alg:evoscore}
\small
\KwIn{seeds $S$ (Table~\ref{tab:seeds}); evolution pool $E$ and validation pool $V$ (disjoint); distribution description $c$; ordered LLM pool $M$; number of registered programs $T$; number of islands $K$; proposals per prompt $s$; re-ranking size $R$}
\KwOut{deployed scoring function}
$f^{*} \leftarrow -\infty$\;
\For(\tcp*[f]{classical warm starts \ding{172}}){$k \leftarrow 1$ \KwTo $K$}{
  $p \leftarrow S[k \bmod |S|]$; $f(p) \leftarrow -$mean strip height of $p$ over $E$\;
  register $p$ in island $k$; $f^{*} \leftarrow \max(f^{*}, f(p))$\;
}
$m \leftarrow$ first model in $M$; $\mathit{failures} \leftarrow 0$; $\mathit{stale} \leftarrow 0$; $t \leftarrow 0$\;
\While{$t < T$}{
  draw an island $k$ uniformly at random\;
  draw up to two elite programs from island $k$ (one while the island holds a single cluster): a cluster of equal-fitness programs is drawn with probability $\propto \operatorname{softmax}(\text{fitness}/\tau)$, then one program from the cluster with preference for shorter code; order the two by fitness\;
  $\mathit{prompt} \leftarrow$ the two programs as \texttt{score\_placement\_v0} and \texttt{\_v1}, the empty header of \texttt{\_v2}, and the description $c$ (\ding{174}, in the system message)\;
  \For(\tcp*[f]{proposals per prompt}){$i \leftarrow 1$ \KwTo $s$ \textbf{while} $t < T$}{
    $p \leftarrow$ body proposed by $m$ for $\mathit{prompt}$, normalized; up to three attempts until $p$ passes the syntax check and the one-instance smoke test\;
    \If{no attempt passed}{
      $\mathit{failures} \leftarrow \mathit{failures} + 1$\;
      \If{$\mathit{failures} \bmod 3 = 0$}{$m \leftarrow$ next model in $M$ \tcp*[f]{model switching \ding{173}}}
      \textbf{continue}\;
    }
    $f(p) \leftarrow -$mean strip height of $p$ over $E$ (one sandboxed, time-limited call); on time-out or crash, discard $p$, $\mathit{stale} \leftarrow \mathit{stale} + 1$ and \textbf{continue}\;
    register $p$ in island $k$; $t \leftarrow t + 1$; $\mathit{failures} \leftarrow 0$\;
    \eIf{$f(p) > f^{*}$}{$f^{*} \leftarrow f(p)$; $\mathit{stale} \leftarrow 0$\;}{$\mathit{stale} \leftarrow \mathit{stale} + 1$\;}
    \If{$\mathit{stale} \ge 15$}{$m \leftarrow$ next model in $M$; $\mathit{stale} \leftarrow 0$ \tcp*[f]{model switching \ding{173}}}
    \If(\tcp*[f]{island reset}){$t \bmod \lfloor T/2 \rfloor = 0$}{
      empty the $\lfloor K/2 \rfloor$ islands with the lowest best fitness and refound each of them with the best program of a surviving island drawn at random\;
    }
  }
}
$\mathcal{C} \leftarrow S \cup \{R \text{ best registered programs by fitness, excluding the seeds}\}$\;
score every program in $\mathcal{C}$ on $V$ with the benchmark simulator\;
\eIf(\tcp*[f]{validation floor \ding{175}}){the best evolved program in $\mathcal{C}$ beats the best seed on $V$}{deploy it\;}{deploy the best seed\;}
\end{algorithm}

The exemplary prompt provided to the LLM, as shown in line 6 of Algorithm~\ref{alg:evoscore}, is detailed in the Appendix in Listing \ref{lst:prompt}.

\section{Experimental design}
\label{sec:design}
\subsection{Overview}\label{sec:overview}
The experimental setup is designed to evaluate the effectiveness, generalization ability, and data dependence of EvoScore for the 2D-OSPP. We analyze two general experimental scenarios:
\begin{enumerate}
    \item synthetic data study, where simulated instances that belong to one of 72 instance classes are used;
    \item logistics data study, where real-world enterprise data are used for generating 2 instance classes (small and large) in two regimes: fitted to real data, and directly including the data, allowing us to investigate how detailed the information about the data is required.
\end{enumerate}

\paragraph{Baselines reference.} In each of these two general scenarios, we compare EvoScore with the reference FunSearch approach, and with the following best-recognized published online heuristic rules:
\begin{itemize}
  \item the five seed heuristics of Table~\ref{tab:seeds}, run unchanged;
  \item the six online algorithms of the open-source rectpack library (MaxRects with the bottom-left, best-short-side-fit, best-area-fit and best-long-side-fit rules; Skyline with the bottom-left and minimum-waste rules), run in true online mode with rotation on their own internal data structures, which makes them a fair external baseline,
   distinctive from our 
   seed heuristics or our method of candidate placement generation.
\end{itemize}
Even though we noticed that some of these 
heuristics (i.e., Max-Contact) perform poorly online, they are included for completeness of the study.
In fact, this clearly shows the
shortcomings of
rules designed primarily for offline use or batch processing.
The methods directly addressing 
the offline settings, such as learned heuristics discussed in
Section~\ref{sub_neural}
are therefore excluded from this evaluation. 

To measure what the four improvements described in Section~\ref{sec:search} add, the original FunSearch approach is run on the same instance classes under the same budget: 100 registered programs over 12 islands, the same evolution, validation and test pools, the same two proposals per prompt and the same token budget per proposal. Following the reference implementation, every island is seeded with the single Best-Fit program. The prompt provides no description of the instance distribution, and the program with the best fitness on the evolution pool is deployed without a validation step. Proposals are drawn from one static model, Deepseek V4 Flash, a member of the EvoScore pool, so the reference differs from EvoScore in exactly the four departures of Section~\ref{sec:search}.

\paragraph{Search configuration.}
The search runs 100 generation steps per instance class over 12 islands, with 100 episodes in the evolution pool, 50 in the validation pool and 100 test episodes, a fixed random seed, $R = 8$ evolved programs entering the final re-ranking, a candidate cap of $C = 40$, and model switching after 3 consecutive malformed or 15 consecutive non-improving proposals.
These hyperparameters were kept fixed throughout the experiments, as we did not observe significant sensitivity to them relative to other factors, such as the data distribution. The main changes with respect to the FunSearch defaults are the island count, raised from 10 to 12 to ensure broader seed coverage with five seeds, and reset and temperature schedules tied to the step count rather than to wall-clock time (Section~\ref{sec:search}).

\subsection{Synthetic data generation}
\label{sec:benchmark}

Synthetic data consists of 72 instance classes.
Each specific instance is not read from a fixed benchmark file, though. Instead, they are sampled from parametric distributions describing each instance class. These characterize a normal distribution, i.e., the mean and standard deviation of item widths 
and aspect ratios (height over width). Both are clipped to fixed allowed ranges (widths to $0.015$--$0.5\,W$, aspect ratios to 0.3--10), and a fixed number of items is drawn per episode. The classes form a full factorial design over four distribution factors and two episode lengths, as shown in Table~\ref{tab:factors}.

\begin{table}[!htbp]
\centering
\caption{The parameters describing considered item dimensions, aspect ratios, and sequence lengths, resulting in 72 distinct classes of synthetic instances.} 
\label{tab:factors}
\small
\begin{tabularx}{0.8\columnwidth}{
    >{\raggedright\arraybackslash}p{0.4\linewidth}
    >{\raggedright\arraybackslash}X
}
\toprule
Factor & Levels \\
\midrule
Aspect ratio mean ($h/w$) & square 1.0, moderate 2.0, elongated 4.0 \\
Aspect ratio std.\ deviation & low 0.1, high 1.0 \\
Item width mean & small $0.075\,W$, medium $0.15\,W$, large $0.3\,W$ \\
Item width std.\ deviation & low $0.015\,W$, high $0.075\,W$ \\
Episode length & 10, 100 items \\
\bottomrule
\end{tabularx}
\end{table}

This design yields $3 \cdot 3 \cdot 2 \cdot 2 \cdot 2 = 72$ classes, each with 100 paired test episodes. We chose a factorial parametric design over a fixed instance library for two reasons. First, EvoScore evolves one heuristic per class, and a class must be a \emph{distribution} from which disjoint evolution, validation and test pools can be drawn, not a finite list of instances.
Second, a factorial design allows us to marginalize over each factor and identify the features that influence the final performance (Section~\ref{sec:results-dims}), an analysis that is intractable with a small number of heterogeneous instance sets.


\subsection{Real-data protocol}
\label{sec:realdata-protocol}

To test EvoScore beyond the parametric sampler, we use a real logistics dataset from the fast-moving consumer goods (FMCG) sector, which is described in more detail in our previous work \cite{ref11}. The dataset consists of a catalog of \textbf{59 distinct rectangular item types}, 
representing goods such as bottled water and dairy products. 
In addition to the catalog of item types, there are about 200 instances of specific packing configurations, i.e., ordered lists of particular items, that represent typical business scenarios. Each instance should be palletized on a strip width $W=1200$~mm, that is, the long side of a Euro pallet. 

Here, we prepare episodes 
based on these real-world data within a two-track experimental setup:
\begin{itemize}
  \item \textbf{Fitted.} A parametric sampler of the same form as in Section~\ref{sec:benchmark} is fitted to the instances distribution from the dataset, as provided in Table~\ref{tab:real_factors}.
  \item \textbf{Directly included.} Episodes draw items independently and uniformly from the catalog. 
\end{itemize}
The resulting fitted synthetic instances, as well as real instances, comprise two classes: small (10 items) and large (100 items). Again, 100 paired  test episodes are used for each instance class.

\begin{table}[!htbp]
\centering
\caption{The parameters describing the distribution of real items' dimensions and aspect ratios, resulting in 2 distinct classes of instances fitted to real data.} 
\label{tab:real_factors}
\small
\begin{tabularx}{0.8\columnwidth}{
    >{\raggedright\arraybackslash}p{0.3\linewidth}
    >{\raggedright\arraybackslash}X
}
\toprule
Factor & Distribution \\
\midrule
Aspect ratio ($h/w$) &  N(0.71, 0.18); range [0.40, 1.0]\\
Item width  &  N($0.196\,W, 0.092\,W$); range [$0.083\,W$, $0.5\,W$] \\
Episode length & two levels: 10, 100 items \\
\bottomrule
\end{tabularx}
\end{table}

Such a design of real case study data protocol allows us to
investigate whether LLM-driven program search requires raw target instances from the user or if statistical summary distributions (fitted parametric models) suffice to achieve competitive performance.
In the latter case, an organization can characterize its item distribution without exposing specific item details to the search loop.



Both regimes use the search configuration of Section~\ref{sec:overview} and the three-pool protocol of Section~\ref{sec:search}. 
Note that in both cases, the prompt context receives the catalog's summary statistics, not the item list.
However, in the fitted case, the evolution and validation pools run on synthetic instances from the sampler, whereas in the real-data case, the instances consist of real items.
On the contrary, the test pool 
used in both fitted and real-data variants 
is the same; that is, it consists of identical real-data instances, so the results can be directly compared.
The fitted regime is additionally evaluated on a synthetic holdout from its own sampler, which separates a shift in instance difficulty from a transfer penalty (Section~\ref{sec:results-real}). 
\section{Results}
\label{sec:results}

\subsection{Ranking and paired comparison}
\label{sec:results-ranking}

\begin{figure*}[!tbp]
\centering
\includegraphics[width=0.5\textwidth]{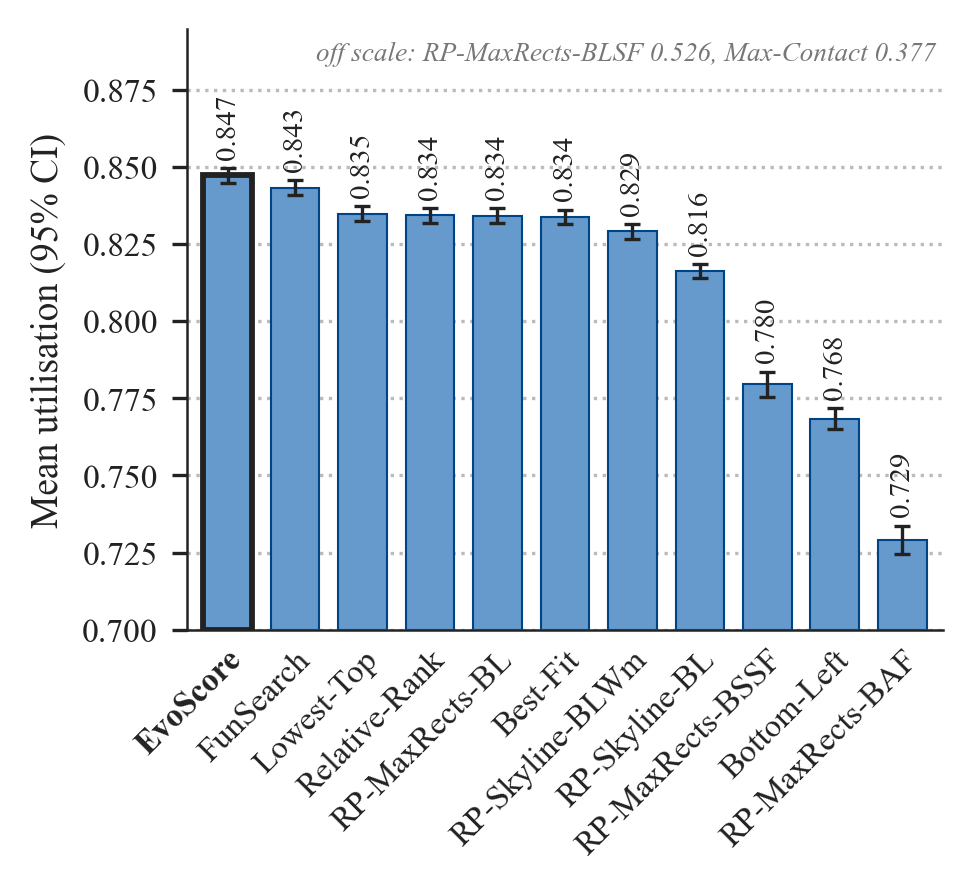}
\caption{Mean utilization per method over all 7,200 synthetic test episodes (bars, value printed above), with 95\% confidence intervals on the mean (whiskers); the methods are sorted, best on the left, and EvoScore is set in bold and outlined. The vertical axis starts at 0.70, and two far weaker methods are listed off scale (RP-MaxRects-BLSF 0.526, Max-Contact 0.377), so that the decisive gaps at the top remain visible.}
\label{fig:ranking}
\end{figure*}

Table~\ref{tab:ranking} reports the mean utilization across the 72 classes $\times$ 100 test episodes, ranked, together with two paired comparisons against EvoScore. The first one is per episode. For each baseline, the column W\,/\,T\,/\,L counts the test episodes, out of 7,200, in which the utilization of EvoScore is higher than (W, win), equal to (T, tie) or lower than (L, loss) that of the baseline on the same instance; $\Delta$ is the mean paired difference (positive means that EvoScore is better), and the last column reports a Wilcoxon signed-rank test on the 7,200 paired differences. The second one, given in the last row, is per class. For each of the 72 classes, the mean utilization of EvoScore is compared with that of the single best competing baseline in that class, and W\,/\,T\,/\,L counts classes. Figure~\ref{fig:ranking} visualizes the ranking.

\begin{table*}[!tbp]
\centering
\caption{Mean utilization over 7,200 test episodes and paired comparison vs.\ EvoScore.}
\label{tab:ranking}
\footnotesize
\setlength{\tabcolsep}{3.2pt}
\begin{tabular}{rlrrrrrlc}
\toprule
Rank & Method & Mean & Median & Q25 & Q75 & $\Delta$ vs.\ EvoScore & L\,/\,T\,/\,W vs.\ EvoScore & $p<0.05$ \\
\midrule
1 & \textbf{EvoScore (ours)} & \textbf{0.8472} & 0.8901 & 0.8010 & 0.9179 & --- & --- (reference) & --- \\
2 & FunSearch & 0.8433 & 0.8843 & 0.7972 & 0.9142 & $+0.0040$ & 2623 / 3052 / 1525 & yes \\
3 & Lowest-Top (MaxRects-BL criterion) & 0.8348 & 0.8767 & 0.7837 & 0.9093 & $+0.0124$ & 3964 / 1791 / 1445 & yes \\
4 & Relative-Rank (MaxRects-BL criterion) & 0.8343 & 0.8742 & 0.7828 & 0.9048 & $+0.0129$ & 3705 / 2405 / 1090 & yes \\
5 & RP-MaxRects-BL & 0.8342 & 0.8744 & 0.7823 & 0.9050 & $+0.0130$ & 3726 / 2401 / 1073 & yes \\
6 & Best-Fit & 0.8337 & 0.8725 & 0.7811 & 0.9045 & $+0.0135$ & 3795 / 2446 / 959 & yes \\
7 & RP-Skyline-BLWm & 0.8290 & 0.8636 & 0.7807 & 0.8972 & $+0.0182$ & 3963 / 2355 / 882 & yes \\
8 & RP-Skyline-BL & 0.8161 & 0.8373 & 0.7749 & 0.8774 & $+0.0311$ & 4295 / 2274 / 631 & yes \\
9 & RP-MaxRects-BSSF & 0.7795 & 0.8532 & 0.7233 & 0.9017 & $+0.0677$ & 5043 / 1297 / 860 & yes \\
10 & Bottom-Left & 0.7684 & 0.8237 & 0.6997 & 0.8723 & $+0.0789$ & 6617 / 133 / 450 & yes \\
11 & RP-MaxRects-BAF & 0.7291 & 0.7998 & 0.6575 & 0.8724 & $+0.1181$ & 6634 / 77 / 489 & yes \\
12 & RP-MaxRects-BLSF & 0.5265 & 0.5560 & 0.3315 & 0.7354 & $+0.3208$ & 7200 / 0 / 0 & yes \\
13 & Max-Contact & 0.3767 & 0.3467 & 0.2112 & 0.4976 & $+0.4706$ & 7135 / 39 / 26 & yes \\
\midrule
 & \textbf{Best competitor per class}$^\dagger$ & 0.8384 & 0.8799 & 0.7863 & 0.9098 & $\mathbf{+0.0089}$ & \textbf{60 / 2 / 10} & yes \\
\bottomrule
\end{tabular}
\begin{flushleft}
\footnotesize $^\dagger$ Per-class head-to-head: for each of the 72 classes, EvoScore is compared against the single best classical baseline in \emph{that} class (the FunSearch reference is not a competitor here; it is compared per class in Section~\ref{sec:results-funsearch}), which gives a win/tie/loss count over \textbf{classes} ($n = 72$) and a per-class mean $\Delta$ of $+0.0089$. All other W\,/\,T\,/\,L counts are per episode ($n = 7{,}200$) against a fixed opponent, so the denominator of this row differs. The RP- prefix denotes the methods of the rectpack library. 

\end{flushleft}
\end{table*}

\begin{figure*}[!tbp]
\centering
\includegraphics[width=\textwidth]{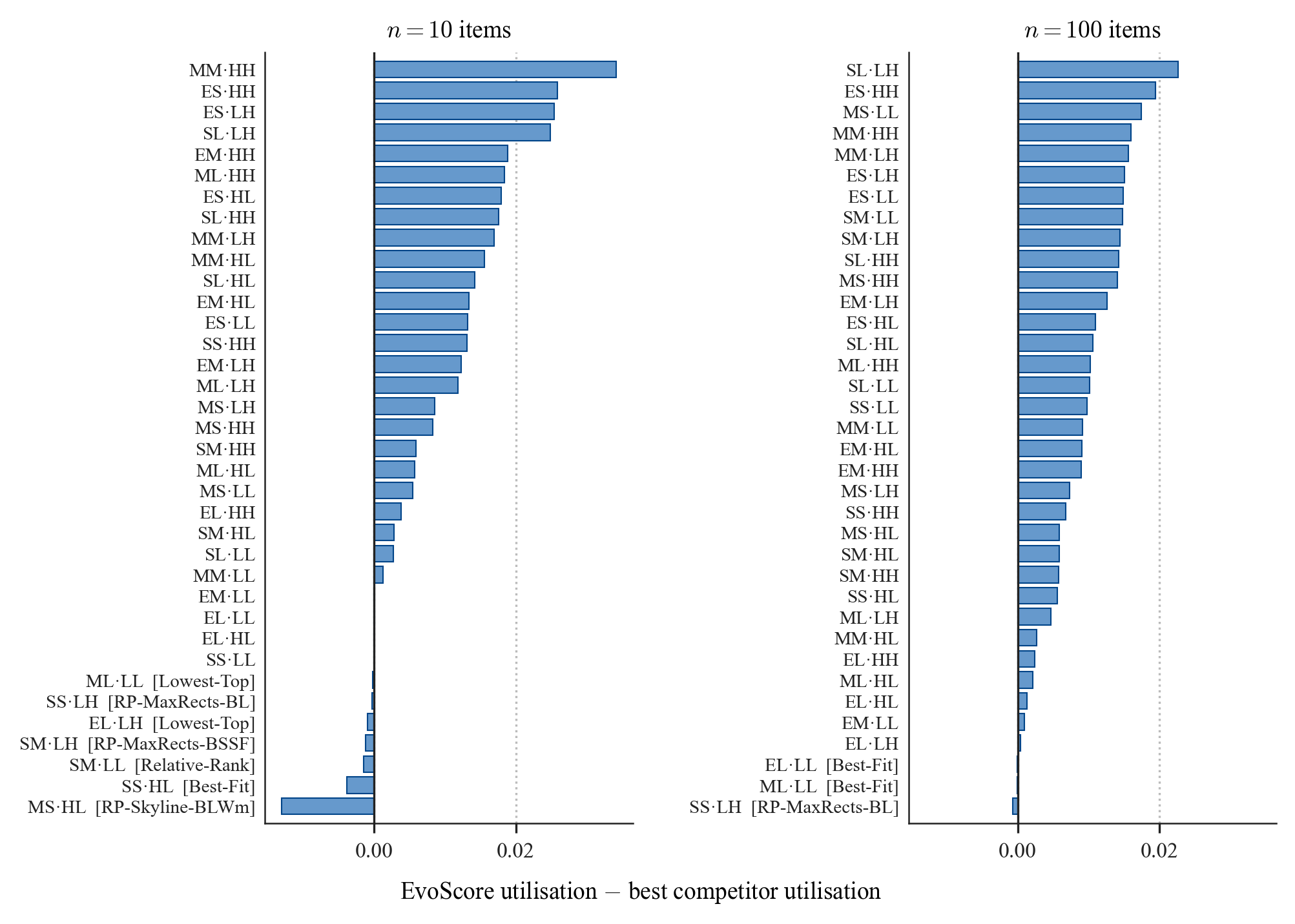}
\caption{Per-class advantage of EvoScore over the best competing baseline in each of the 72 classes, sorted within episode length (left $n = 10$, right $n = 100$). Bars to the right of zero are classes won by EvoScore, bars to the left are losses; square brackets give the winning baseline. Class codes give one letter per design factor of Table~\ref{tab:factors}, in the order mean aspect ratio (E elongated, M moderate, S square), mean item width (S small, M medium, L large), aspect-ratio standard deviation (H high, L low) and width standard deviation (H high, L low); for example EL$\cdot$HL is an elongated, large-item class with high aspect-ratio variance and low width variance. EvoScore wins 60, ties 2 and loses 10 classes.}
\label{fig:advantage}
\end{figure*}

EvoScore is the top method in the suite. It wins with the next best method by 1.24 percentage points  (0.8472 vs.\ 0.8348), outperforming every online baseline on a per-episode basis, and wins 60 of 72 classes outright against the best competitor in each (Figure~\ref{fig:advantage}). The method-to-method comparison is intentionally conservative, namely the opponent is selected post hoc against EvoScore in every class, so the 60/2/10 result is a lower bound on the dominance relative to any single fixed baseline. The per-class margin over the best baseline heuristic is smaller but still statistically significant, 0.89 percentage points on average. With 100 episodes per class, the median 95\% confidence interval of a per-class mean has a half-width of 0.4 percentage points.

\subsection{Performance by instance dimension}
\label{sec:results-dims}
\begin{figure*}[!tbp]
\centering
\includegraphics[width=0.5\textwidth]{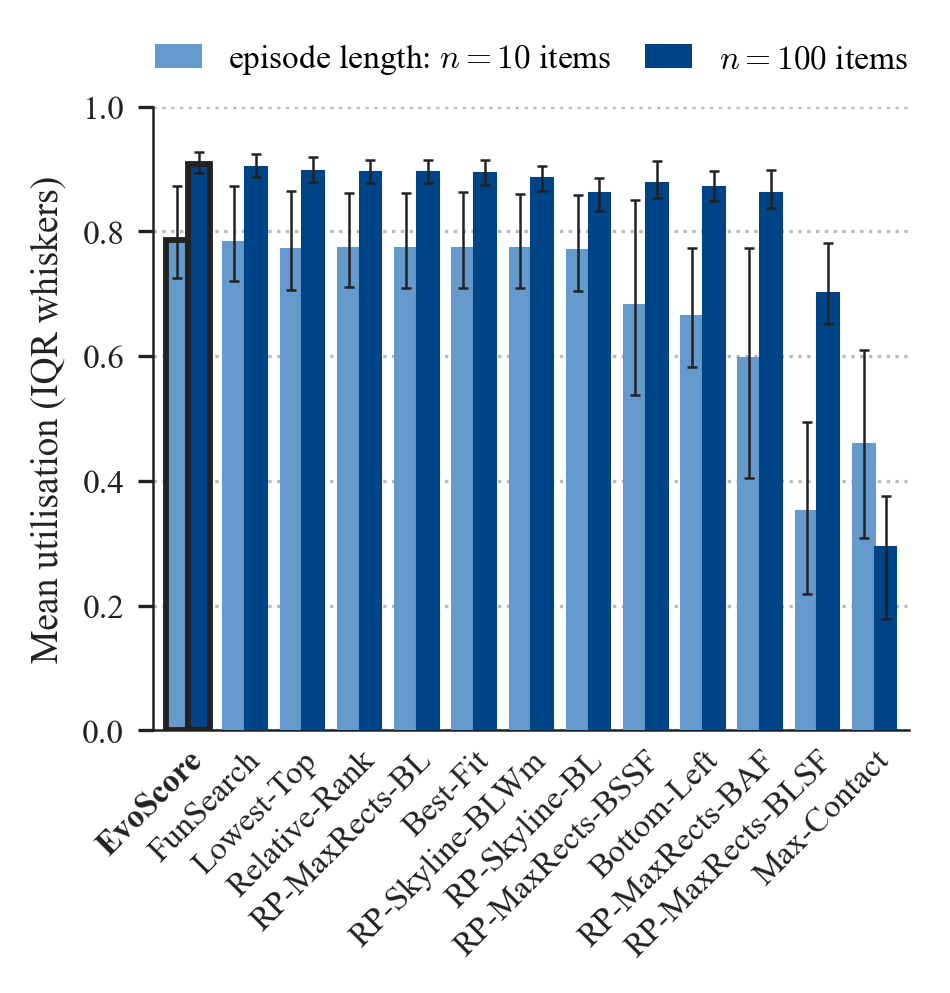}
\caption{Mean utilization by episode length per method: the light bar is the 10-item mean and the dark bar the 100-item mean, each with interquartile whiskers (the methods are ordered by the overall mean, as in Figure~\ref{fig:ranking}; EvoScore is set in bold and outlined).}
\label{fig:episode-length}
\end{figure*}

The aggregate ranking does not reveal the conditions under which the advantage arises. To disentangle the contributing factors, the results are marginalized over the factorial design described in Table \ref{tab:factors}: for each factor, one level is held fixed and the utilization is averaged over all classes sharing that level, so that each level comprises 24 or 36 classes with 100 episodes each. ``Best baseline'' denotes the strongest non-EvoScore method at that level (marginal mean), and ``EvoScore wins'' counts the classes at that level in which EvoScore beats the best competitor in that class. The results are presented in Table~\ref{tab:marginal}; Figure~\ref{fig:episode-length} shows the breakdown by episode length.

\begin{table*}[!tbp]
\centering
\caption{Marginal performance by design factor.}
\label{tab:marginal}
\footnotesize
\setlength{\tabcolsep}{3.5pt}
\begin{tabular}{llrrlrrr}
\toprule
Dimension & Level & \#classes & EvoScore & Best baseline (marginal) & $\Delta$ vs.\ best & EvoScore wins & mean $\Delta$/class \\
\midrule
\multirow{3}{*}{Mean aspect ratio ($h/w$)} & square & 24 & 0.8054 & 0.7927 (RP-MaxRects-BL) & $+0.0127$ & 18/24 & $+0.0081$ \\
 & moderate & 24 & 0.8474 & 0.8356 (Lowest-Top) & $+0.0118$ & 21/24 & $+0.0091$ \\
 & elongated & 24 & 0.8889 & 0.8776 (Lowest-Top) & $+0.0112$ & 21/24 & $+0.0094$ \\
\midrule
\multirow{3}{*}{Mean item width} & small & 24 & 0.7969 & 0.7846 (Best-Fit) & $+0.0123$ & 19/24 & $+0.0095$ \\
 & medium & 24 & 0.8532 & 0.8417 (Lowest-Top) & $+0.0115$ & 22/24 & $+0.0098$ \\
 & large & 24 & 0.8916 & 0.8835 (Lowest-Top) & $+0.0081$ & 19/24 & $+0.0074$ \\
\midrule
\multirow{2}{*}{Aspect-ratio std.\ dev.} & low & 36 & 0.8479 & 0.8372 (Lowest-Top) & $+0.0107$ & 27/36 & $+0.0080$ \\
 & high & 36 & 0.8465 & 0.8325  (Best-Fit) & $+0.0141$ & 33/36 & $+0.0097$ \\
\midrule
\multirow{2}{*}{Width std.\ dev.} & low & 36 & 0.8511 & 0.8412 (Best-Fit) & $+0.0099$ & 28/36 & $+0.0057$ \\
 & high & 36 & 0.8434 & 0.8299 (Lowest-Top) & $+0.0135$ & 32/36 & $+0.0120$ \\
\midrule
\multirow{2}{*}{Episode length} & 10 items & 36 & 0.7857 &  0.7739 (Relative-Rank) & $+0.0118$ & 27/36 & $+0.0088$ \\
 & 100 items & 36 & 0.9087 & 0.8970 (Lowest-Top) & $+0.0117$ & 33/36 & $+0.0089$ \\
\bottomrule
\end{tabular}
\end{table*}

Three patterns can be observed from the marginalized results. First, EvoScore leads at every level of every factor. The absolute utilization of every method rises with elongation ($0.805 \rightarrow 0.889$ for EvoScore, $0.793 \rightarrow 0.878$ for the best baseline), item size ($0.797 \rightarrow 0.892$) and episode length ($0.786 \rightarrow 0.909$). This rise is a property of the instances, because elongated items of similar width stack into columns with little gap between them, large items fill the strip width with fewer pieces and leave less fragmented free space, and long episodes spread the waste of the ragged top edge over more items. Relative to the best marginal baseline, the advantage declines from $+0.0127$ for square to $+0.0112$ for elongated items and from $+0.0123$ for small to $+0.0081$ for large items. The per-class mean follows this decline for item width ($+0.0095 \rightarrow +0.0074$) but rises slightly for the aspect ratio ($+0.0081 \rightarrow +0.0094$). The interpretation is that dense regimes leave less room for improvement. At a utilization near 0.89, the remaining waste is small and largely unavoidable for an online policy, and large items in particular admit few alternative positions, so that fewer ties remain for an evolved tie-break to resolve. Second, the advantage grows with heterogeneity. Under high aspect-ratio variance, the edge over the best baseline widens to $+0.0141$ (33 of 36 classes won), and under high width variance to $+0.0135$ (32 of 36, $+0.0120$ mean per class), versus $+0.0107$ and $+0.0099$ at the corresponding low-variance levels. Third, EvoScore wins most consistently on long episodes (33/36 at $n = 100$ vs.\ 27/36 at $n = 10$). All three patterns are consistent with the mechanism identified in Section~\ref{sec:results-seeds}, namely the gains come from tie-breaking among many candidate placements, and heterogeneous, longer streams create both more ties to break and more room to break them well. The thinnest edge is observed for large items ($+0.0081$), where a few big rectangles leave little freedom for tie-break refinement.

\subsection{How the evolved heuristics differ from their seeds}
\label{sec:results-seeds}

Under the validation-floor protocol, \textbf{68 of 72 classes retained a genuinely evolved heuristic}. Only four classes reverted to a seed because no evolved program beat the seed on held-out instances (three Best-Fit, one Lowest-Top). Interestingly, the form of the improvement is consistent across classes. Evolution keeps the primary objective and the lexicographic hierarchy of the seed unchanged and adds one to three mid-weight terms that refine the tie-breaking among placements that are equal on the primary objective. In other words, the result is interpretable, additive tie-break engineering on a classical skeleton, not a restructuring of the decision rule.

\begin{figure*}[!tbp]
\centering
\includegraphics[width=\textwidth]{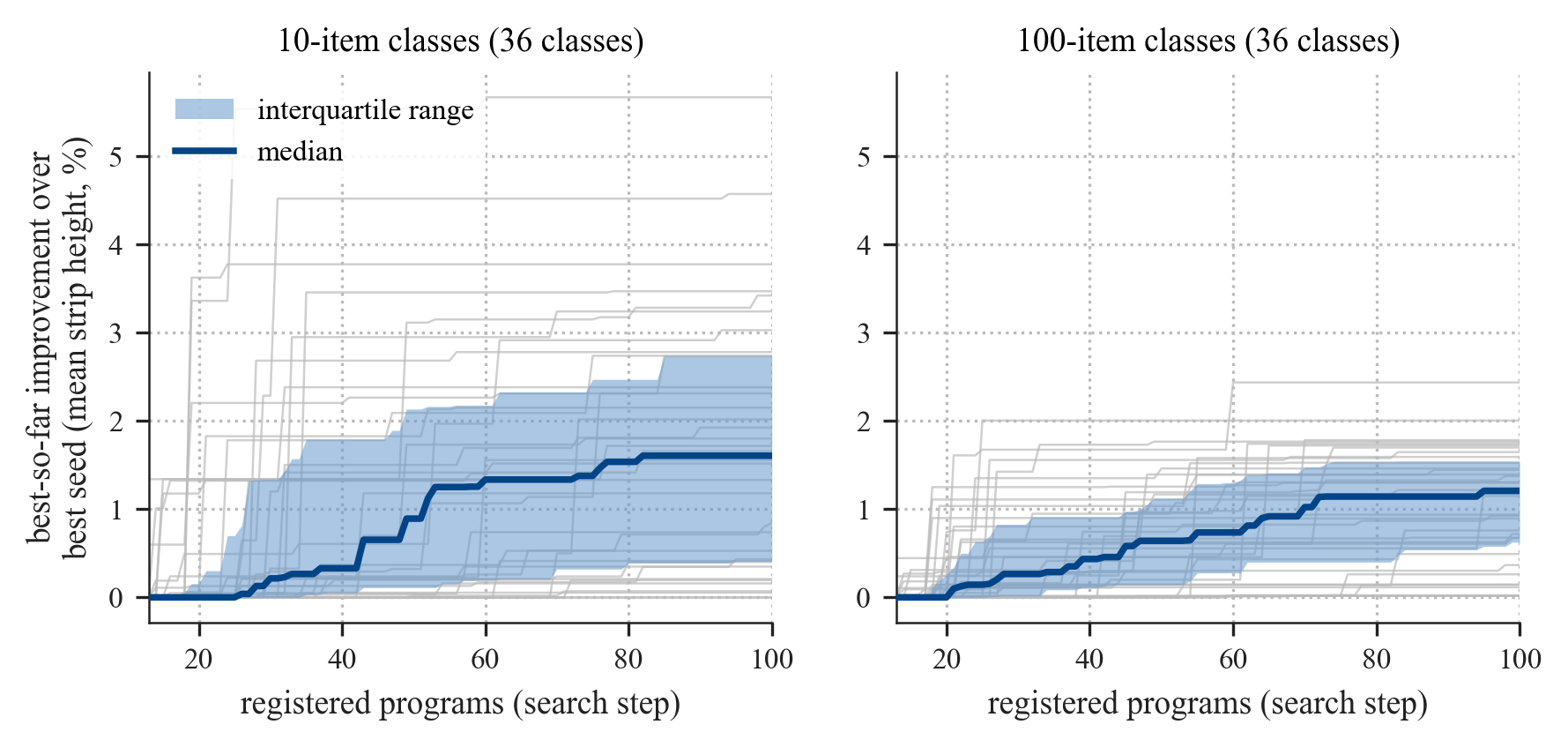}
\caption{Convergence of the search: best-so-far fitness on the evolution pool, expressed as the relative reduction of the mean strip height with respect to the best seed, against the search step (number of registered programs; the first twelve are the seeds), for the 36 ten-item classes (left) and the 36 hundred-item classes (right). Thin gray lines are individual classes, the thick dark line is the median, and the light band is the interquartile range.}
\label{fig:convergence}
\end{figure*}

Figure~\ref{fig:convergence} shows how the search converges. For every class, the best-so-far fitness on the evolution pool is plotted against the search step and expressed as the relative reduction of the mean strip height with respect to the best seed. The median reduction at the end of the search is 1.6\% for the 10-item classes and 1.2\% for the 100-item classes (mean over all classes 1.4\%). The median class reaches 90\% of its final improvement at step 60 (upper quartile at step 73), and the median curve itself crosses 90\% of its final value at step 72. The improvement is step-like, as expected of a best-so-far statistic. The curves flatten over the second half of the search, 13 of the 72 classes still improved in the last ten steps, and in two classes no evolved program beat the best seed on the evolution pool.

The evolved programs grow from the 1 to 23 lines of the seeds to a median of 48 lines. The recurring added terms (counts out of 68) are wall or edge affinity (51), gap, waste or neighbor clustering (50), contact or overlap (34), and support from below or corner anchoring (33). Four programs are examined in detail below. For each one, the mean test utilization of the evolved program in its class is compared with that of its seed, and the lines that change the decision rule are quoted; the complete programs are reproduced in Appendix~\ref{app:programs}.
\begin{enumerate}
  \item \textbf{Lowest-Top lineage} (moderate aspect ratio, medium width, high variance of both, 10 items): 0.7924 vs.\ seed 0.7576 ($+3.5$\,pp). The primary term of the seed, the normalized top edge, is retained with its weight. The tie-break $-x$ of the seed is replaced by a penalty on the distance to the nearer side wall, and the $x$ term is reduced to a third level:
\begin{lstlisting}
side_penalty = -min(left_space, right_space) * 0.1
height_penalty = -(top - min_top) / span * 10000
x_penalty = -candidate["x"] * 0.01
return height_penalty + side_penalty + x_penalty
\end{lstlisting}
  Among placements with the same top edge, the candidate adjacent to a side wall is therefore selected, which prevents narrow unused regions on both sides of an item.
  \item \textbf{Best-Fit lineage} (square, large, high aspect-ratio variance, low width variance, 10 items): 0.8128 vs.\ 0.7958 ($+1.7$\,pp). The Best-Fit core and its weights are unchanged. A discrete bonus of 1000 is added for candidates that touch a side wall and are supported from below, by the floor or by a placed item:
\begin{lstlisting}
corner_score = 0
if (touches_left_wall or touches_right_wall) and (touches_bottom or touches_placed_item_below):
  corner_score = 1000
return -(height_increase * 10000 + candidate["y"] * 100 + candidate["x"] - corner_score)
\end{lstlisting}
  The bonus outranks the $x$ term but not the $y$ term, so that the order of criteria becomes height increase, vertical position, supported corner, horizontal position.
  \item \textbf{Best-Fit lineage, long episodes} (square, large, low aspect-ratio variance, high width variance, 100 items): 0.8900 vs.\ 0.8546 ($+3.5$\,pp). The height-increase term remains dominant. Three terms that act within a row are added: a penalty on the margin to the right wall, a bonus for touching either wall, and a bonus for a small horizontal gap to a placed item in the same vertical band:
\begin{lstlisting}
cluster_bonus = -x_gap * 2 if x_gap < 20 else 0
score = -height_increase * 10000 - y * 50 - x * 5 - right_margin * 3 + edge_bonus * 10 + cluster_bonus
\end{lstlisting}
  These terms improve the horizontal compaction of the rows, which matters on long episodes. The program also computes a variable \texttt{gap\_penalty} that does not enter the score; such inactive code occurs in LLM-generated programs and has no effect on the decision.
  \item \textbf{Max-Contact lineage} (moderate, medium, high aspect-ratio variance, low width variance, 10 items): 0.7984 vs.\ seed 0.3614 ($+43.7$\,pp). The seed maximizes the contact length and contains no term that depends on the strip height, therefore it builds tall stacks and achieves a poo utilization of 0.3614. The evolved program retains the complete contact computation and adds two terms, a penalty proportional to the packing height after the placement and a fixed penalty for candidates that are not supported from below:
\begin{lstlisting}
score = contact - 5.0 * delta_height
# ... has_bottom_contact: a placed item ends at cy and overlaps the candidate in x
if not has_bottom_contact:
  score -= 50.0
\end{lstlisting}
  The two added terms supply the objective that the seed lacks, and the resulting program is the best method in its class. The search can thus introduce a missing objective and not only adjust the weights of existing terms.
\end{enumerate}

In summary, the discovered programs are not new packing paradigms but interpretable additive tie-break engineering on classical skeletons, which is sufficient to beat every online baseline in our suite.

\subsection{Results on real logistics data}
\label{sec:results-real}

\begin{figure*}[!tbp]
\centering
\includegraphics[width=0.5\textwidth]{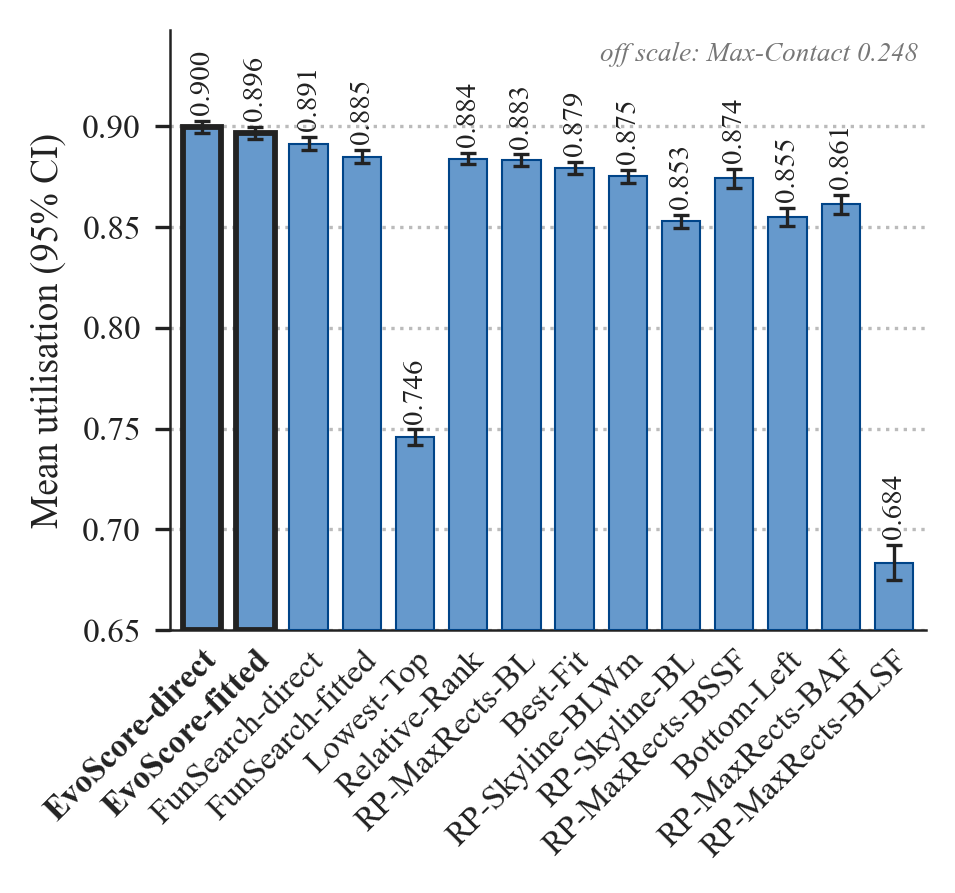}
\caption{Mean utilization per method on the real logistics holdout, $n = 100$ (bars, value printed above), with 95\% confidence intervals on the mean (whiskers); the methods follow the order of Figure~\ref{fig:ranking}, with EvoScore-direct and EvoScore-fitted in EvoScore's place (bold, outlined). The two EvoScore variants occupy the top two ranks, ahead of both FunSearch variants; Max-Contact (0.248) is listed off scale.}

\label{fig:real-ranking}
\end{figure*}

\begin{figure*}[!tbp]
\centering
\includegraphics[width=\textwidth]{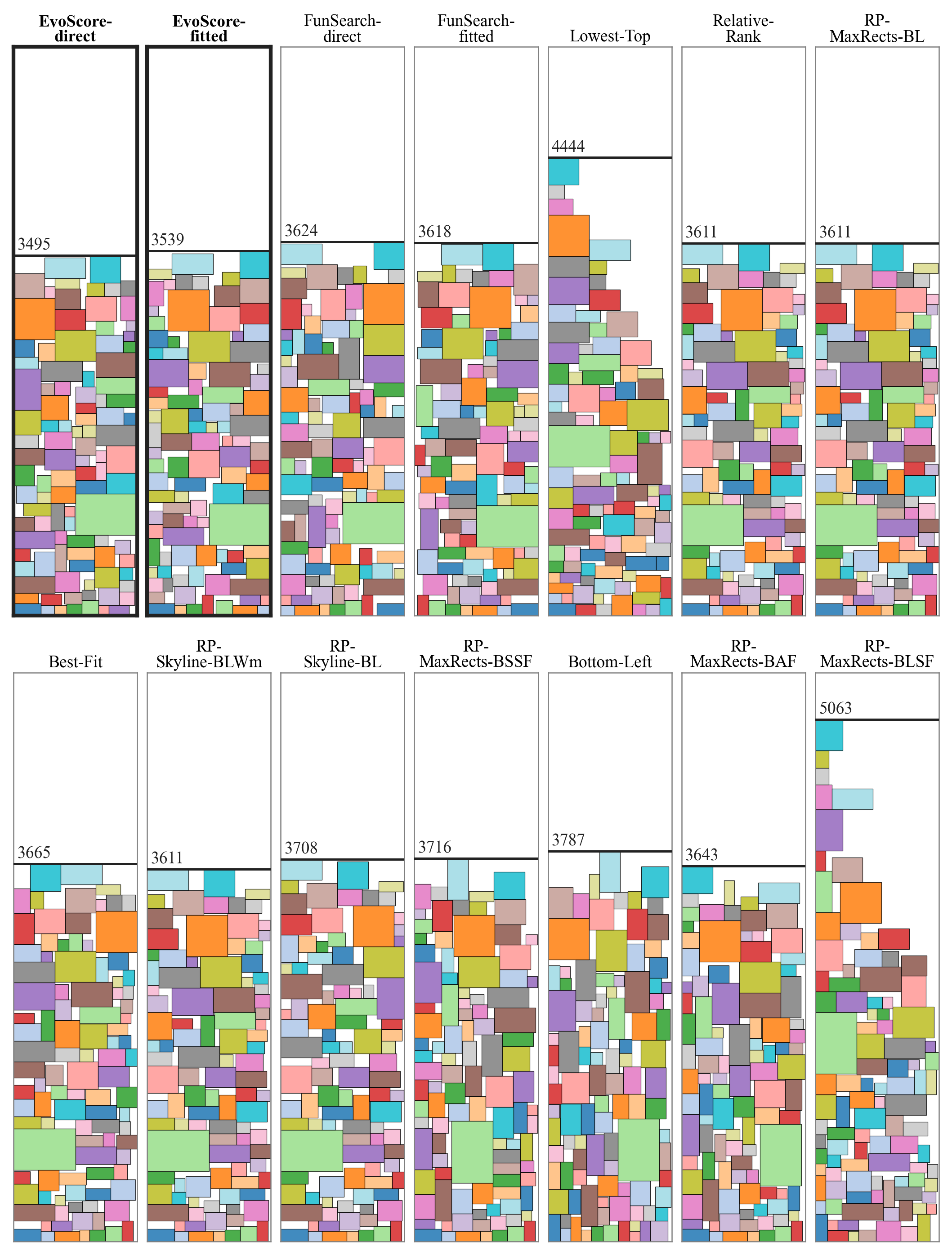}
\caption{One real-holdout episode ($n = 100$) packed by every method (shared vertical scale; a horizontal line marks each realized strip height, printed in grid units of $1/1200\,W$; Max-Contact exceeds the scale and is labeled with its height). The two EvoScore panels are outlined and titled in bold. EvoScore-direct attains the lowest strip height ($2.91\,W$, utilization 0.901) and EvoScore-fitted the second lowest among the candidate-scoring methods ($2.95\,W$, 0.890).}
\label{fig:real-packings}
\end{figure*}

Table~\ref{tab:real} reports the mean utilization on the shared real holdout (100 paired episodes per episode length; protocol of Section~\ref{sec:realdata-protocol}). Figure~\ref{fig:real-ranking} visualizes the $n = 100$ block, and Figure~\ref{fig:real-packings} shows one holdout episode packed by every method except Max-Contact, whose strip height is more than three times that of the others.

\begin{table}[!htbp]
\centering
\caption{Mean utilization on the real logistics holdout (rows sorted by the $n = 100$ column).}
\label{tab:real}
\small
\begin{tabular}{lrr}
\toprule
Method & $n = 10$ & $n = 100$ \\
\midrule
\textbf{EvoScore-direct (ours)} & \textbf{0.7406} & \textbf{0.8997} \\
\textbf{EvoScore-fitted (ours)} & 0.7269 & \textbf{0.8964} \\
FunSearch-direct & 0.7282 & 0.8913 \\
FunSearch-fitted & 0.7215 & 0.8847 \\
Relative-Rank & 0.7317 & 0.8838 \\
RP-MaxRects-BL & 0.7315 & 0.8830 \\
Best-Fit & 0.7228 & 0.8793 \\
RP-Skyline-BLWm & 0.7303 & 0.8751 \\
RP-MaxRects-BSSF & 0.6481 & 0.8740 \\
RP-MaxRects-BAF & 0.6215 & 0.8612 \\
Bottom-Left & 0.6317 & 0.8548 \\
RP-Skyline-BL & 0.7283 & 0.8527 \\
Lowest-Top & 0.6889 & 0.7458 \\
RP-MaxRects-BLSF & 0.2819 & 0.6836 \\
Max-Contact & 0.4368 & 0.2476 \\
\bottomrule
\end{tabular}
\end{table}

Based on the results presented in the Table \ref{tab:real}, there are four main conclusions. 
First, EvoScore is able to transfer to real data. At $n = 100$, both variants occupy the top two ranks and significantly beat every baseline under the per-episode paired Wilcoxon test (direct vs.\ best baseline: $\Delta = +0.0158$, 85/1/14 W\,/\,T\,/\,L, $p = 2.1 \times 10^{-13}$; fitted: $\Delta = +0.0125$, 77/0/23, $p = 5.9 \times 10^{-10}$). At $n = 10$, EvoScore-direct achieves the highest mean utilization (0.7406), but its advantage over the best baseline, Relative-Rank, is not statistically significant ($\Delta = +0.0089$, 22/55/23 W\,/\,T\,/\,L, $p = 0.25$): on short episodes, 55 of the 100 paired episodes end in exactly the same utilization, so that the paired test has little power. EvoScore-fitted (0.7269) falls 0.005 below the best baseline, a difference that is likewise not significant (23/39/38, $p = 0.33$). The six methods between 0.7269 and 0.7406 thus form a group whose internal ranking the 100 holdout episodes cannot resolve; EvoScore-direct leads this group and EvoScore-fitted closes it.

\begin{figure}[!htbp]
\centering
\includegraphics[width=0.5\textwidth]{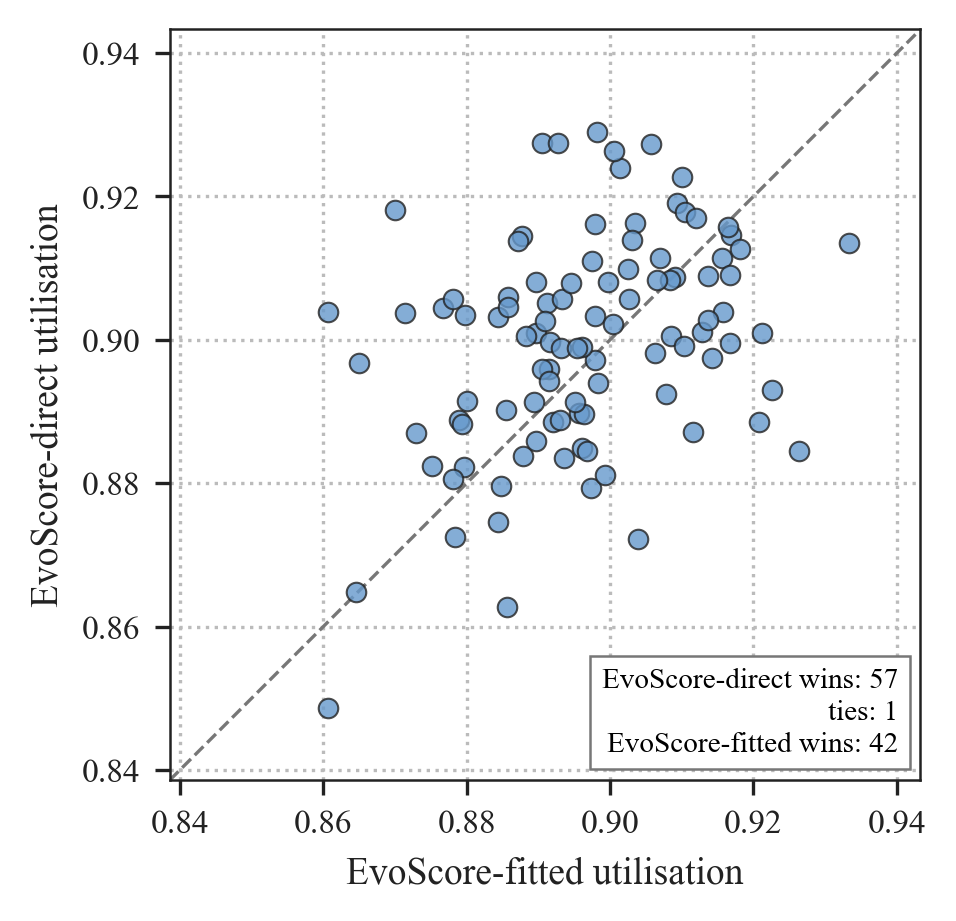}
\caption{Paired per-episode comparison of EvoScore-direct vs.\ EvoScore-fitted on the byte-identical real holdout, $n = 100$. The points scatter tightly around the diagonal (57 wins / 1 tie / 42 losses; mean $\Delta = +0.0033$: evolving against fitted distribution parameters recovers essentially the full benefit of evolving on the real items.}
\label{fig:direct-vs-fitted}
\end{figure}

Second, distribution knowledge can be sufficient to evolve effective programs using EvoScore. On identical instances, the advantage of the direct variant over the fitted one at $n = 100$ is $+0.0033$ (57/1/42 W\,/\,T\,/\,L; Wilcoxon $p = 0.062$). At $n = 10$, the two variants are not equivalent: the direct variant is genuinely better ($+0.0137$, 36/47/17, $p = 0.004$. In practical terms, the search does not need access to the real items. A heuristic evolved on synthetic instances drawn from a distribution fitted to the catalogue, which requires only the mean and the standard deviation of the item width and of the aspect ratio together with their observed bounds, achieves on long episodes the same utilization as a heuristic evolved on the real items themselves, within one percentage point. Access to the real instances pays off only on short episodes, where the outcome of a ten-item packing depends on the particular items drawn, and such instance-level detail is not captured by the fitted distribution (Figure~\ref{fig:direct-vs-fitted}).

\begin{figure}[!tbp]
\centering
\includegraphics[width=0.5\textwidth]{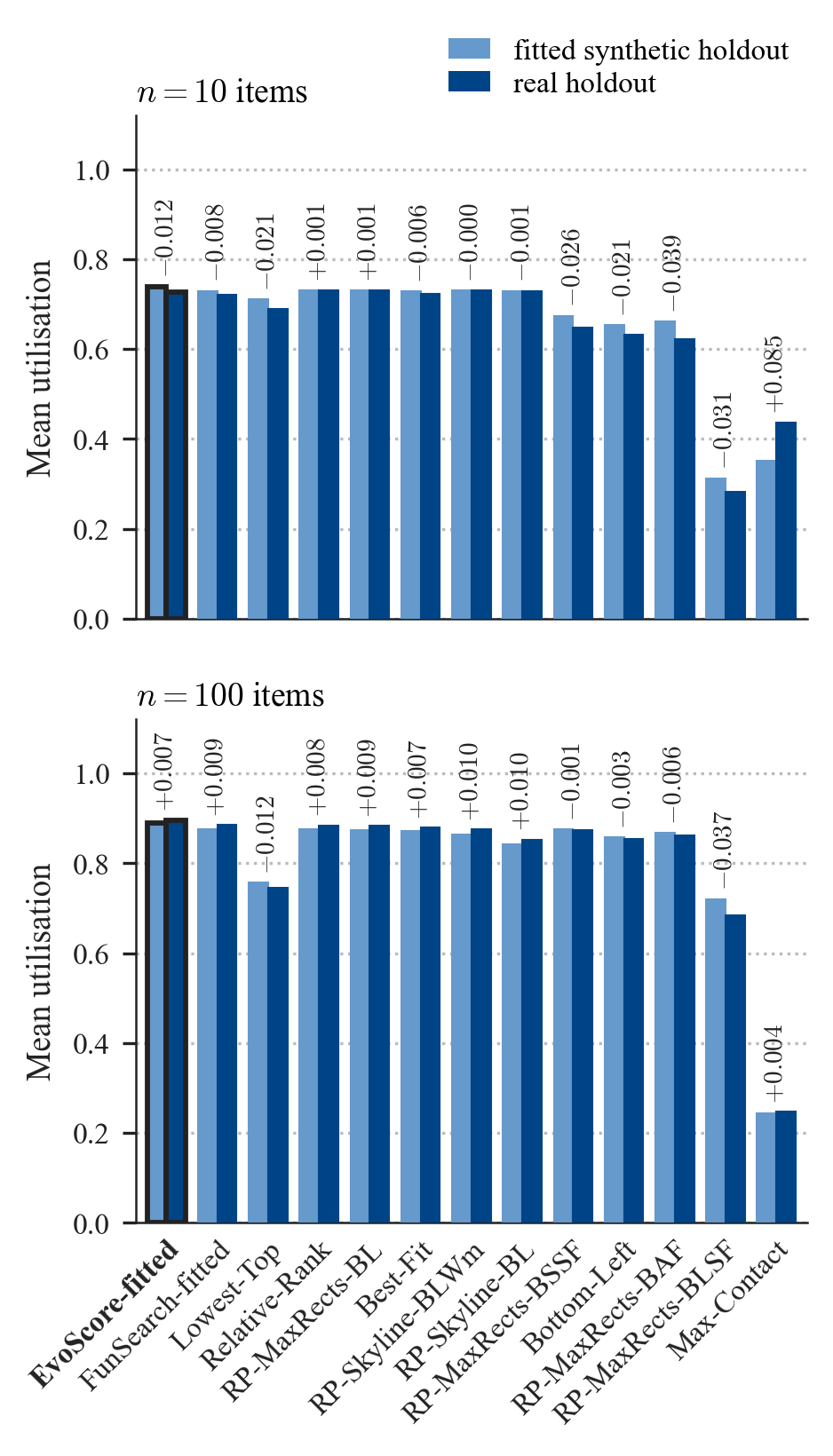}
\caption{Generalization of the fitted setting: mean utilization on the fitted synthetic holdout (light bars) vs.\ the real holdout (dark bars), per method and episode length (top $n = 10$, bottom $n = 100$; the methods follow the order of Figure~\ref{fig:ranking}, with FunSearch-fitted next to EvoScore-fitted; the labels give the signed difference, real minus synthetic). EvoScore-fitted moves within the same band as the distribution-agnostic baselines, i.e., there is no distribution-shift cliff.}
\label{fig:generalization}
\end{figure}

Third, there is no distribution-shift cliff. When the synthetic holdout of the fitted variant (its own training distribution) is compared with the real holdout, its utilization moves by $+0.007$ at $n = 100$, which is inside the $+0.007$ to $+0.010$ band by which the distribution-agnostic baselines also move (the discrete catalogue packs slightly denser than the fitted continuous model), and by $-0.012$ at $n = 10$, a modest transfer cost where the strongest baselines are flat (Figure~\ref{fig:generalization}).

Fourth, the pipeline improvements matter more on real data. The FunSearch reference of Section~\ref{sec:benchmark}, run in both regimes on the same pools and budget, reaches 0.8913 (direct) and 0.8847 (fitted) at $n = 100$ (Table~\ref{tab:real}, Figure~\ref{fig:real-ranking}). On the paired holdout, EvoScore-direct beats FunSearch-direct by $+0.0084$ (64/1/35 W\,/\,T\,/\,L, $p = 6.8 \times 10^{-6}$) and EvoScore-fitted beats FunSearch-fitted by $+0.0117$ (80/0/20, $p = 7.9 \times 10^{-10}$), two to three times the 0.40-point margin of the synthetic benchmark (Section~\ref{sec:results-funsearch}). The gap is widest in the fitted regime, which is the deployment scenario: FunSearch-fitted is statistically indistinguishable from the best baseline ($+0.0009$ vs.\ Relative-Rank, 54/1/45, $p = 0.47$), while EvoScore-fitted keeps an edge of $+0.0125$ over the same baseline. The direct versus fitted gap of FunSearch is $+0.0066$ (62/1/37, $p = 2.6 \times 10^{-4}$), twice the $+0.0033$ of EvoScore, so the equivalence of the second finding is a property of the full pipeline, not of program search as such. At $n = 10$, EvoScore-direct beats FunSearch-direct by $+0.0124$ (29/49/22, $p = 0.044$), the two fitted variants do not differ significantly ($+0.0054$, $p = 0.12$), and FunSearch-fitted falls significantly below the best baseline ($-0.0102$, $p = 0.004$): the fitness arg-max deployed a program whose validation score was below that of Relative-Rank, precisely the case the validation floor exists for.

The discovered programs continue the pattern of Section~\ref{sec:results-seeds}, with one addition worth noting. The fitted $n = 100$ program (Listing~\ref{lst:ex-e}) keeps the Best-Fit skeleton and adds a wall-hugging minimum-side-gap term plus an orientation term that penalizes the absolute deviation of the aspect ratio of the candidate from 0.71:
\begin{lstlisting}
aspect_ratio_deviation = abs(aspect_ratio - 0.71)
quaternary_score_ar = -aspect_ratio_deviation * 100
\end{lstlisting}
The constant 0.71 is exactly the fitted mean aspect ratio stated in its prompt context. Context-conditioned prompting (Section~\ref{sec:search}, \ding{174}) thus demonstrably shapes the discovered rule itself, and not merely its fitness. The direct $n = 100$ program (Listing~\ref{lst:ex-f}) goes beyond tie-break editing. It reconstructs the skyline from the placed items, counts the length of the bottom edge of the candidate that rests on the skyline, and weighs this supported length against the top edge of the candidate and its horizontal position:
\begin{lstlisting}
height_penalty = top * 1000
left_bonus = -candidate["x"] * 10
support_bonus = supported_perimeter * 500
return support_bonus - height_penalty + left_bonus
\end{lstlisting}
The program thus re-derives a support/adhesion trade-off from the data alone, and it is the strongest evolved program in the study. Both programs are presented in Appendix~\ref{app:programs} (Listings~\ref{lst:ex-e} and~\ref{lst:ex-f}). 

\subsection{Comparison with the FunSearch recipe}
\label{sec:results-funsearch}

\begin{figure*}[!tbp]
\centering
\includegraphics[width=\textwidth]{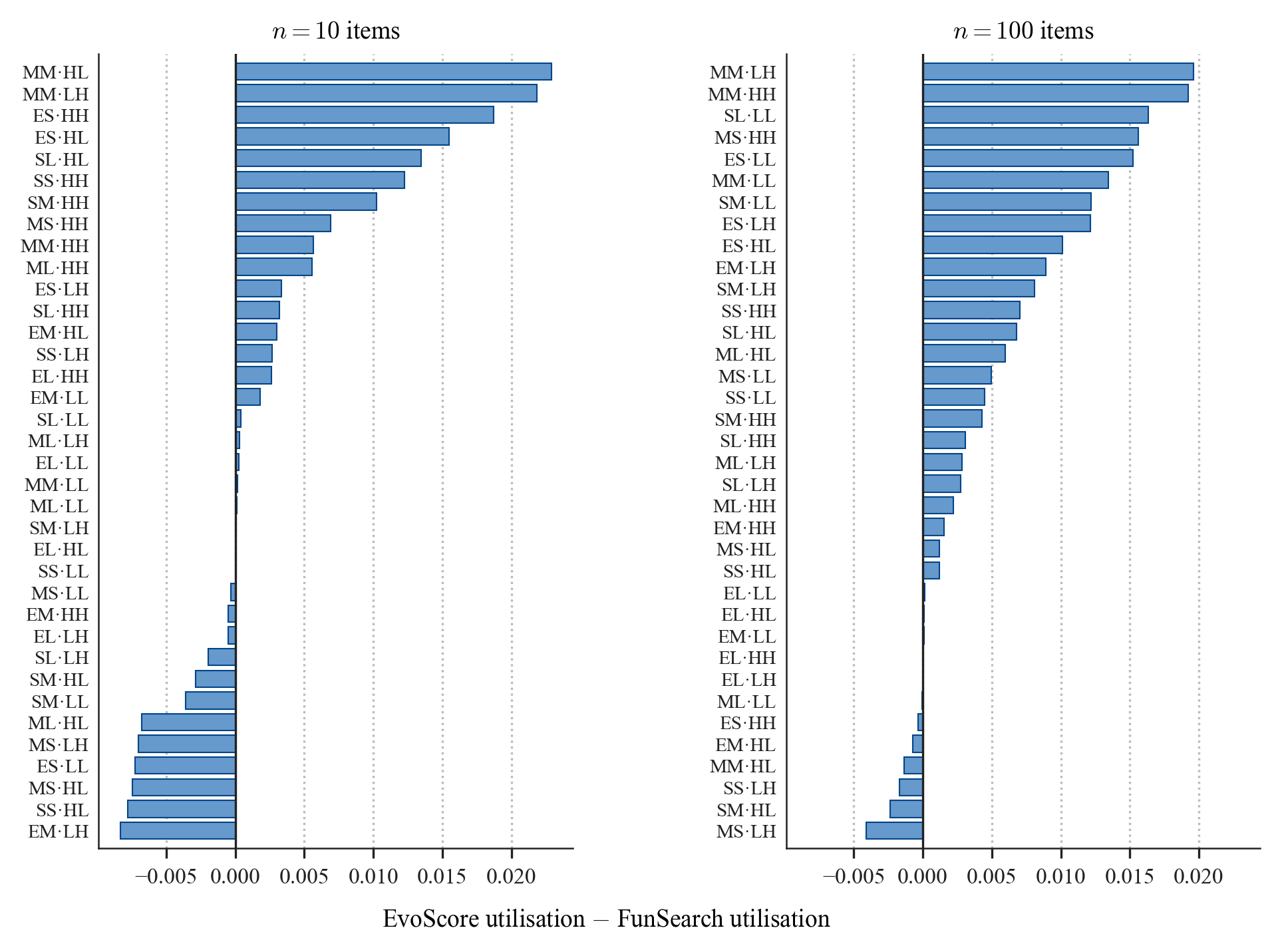}
\caption{Per-class difference in mean utilization between EvoScore and the FunSearch reference, sorted within episode length (left $n = 10$, right $n = 100$); class codes as in Figure~\ref{fig:advantage}. Bars to the right of zero are classes won by EvoScore. EvoScore wins 50, ties 2 and loses 20 classes; the largest loss is 0.8 points and the largest win 2.3 points.}
\label{fig:funsearch}
\end{figure*}

The four modifications described in Section~\ref{sec:search} are measured jointly against the FunSearch reference of Section~\ref{sec:benchmark}, which runs the original method with DeepSeek V4 Flash on the same 72 classes, pools and budget. The reference is the second row of Table~\ref{tab:ranking}, and Figure~\ref{fig:funsearch} shows the per-class differences.

EvoScore gains 0.40 percentage points of utilization over the FunSearch recipe (0.8472 against 0.8433; Wilcoxon $p = 2.4 \cdot 10^{-61}$), a relative improvement of 0.5\%: EvoScore wins 2,623 of the 7,200 test episodes, loses 1,525 and ties the remaining 3,052, and it wins 50 of the 72 classes against 20 losses. The FunSearch recipe is itself a strong method. It ranks second in Table~\ref{tab:ranking} and Figure~\ref{fig:ranking}, ahead of every classical rule (0.8433 against 0.8348 for Lowest-Top), and it beats the best classical rule in 51 of 72 classes, by 0.49 percentage points per class on average; EvoScore beats the same opponent in 60 of 72 classes by 0.89 points. The four departures together thus account for close to half of the margin of EvoScore over the classical heuristics. The gap is wider on long episodes ($+0.53$ points, 0.9087 against 0.9035, 28 of 36 classes won) than on short ones ($+0.27$ points, 0.7857 against 0.7830, 22 of 36 classes won, 2 tied), the pattern already seen in Section~\ref{sec:results-dims}: long streams create more ties to break, and the departures pay off where the evolved tie-break has more to decide.

Figure~\ref{fig:funsearch} shows where the two methods part. The 20 classes won by FunSearch are small-margin cases, 12 of them ten-item classes, and none is lost by more than 0.8 points, whereas the wins of EvoScore reach 2.3 points and concentrate in the heterogeneous medium-width classes. The reference also lacks the deployment guarantee of the validation floor: it deploys the fitness-maximizing program of every class, and in 19 of the 72 classes that program is below the best classical rule on the test pool, against 10 such classes for EvoScore. On the real logistics data, where the reference is run in both regimes, the margin is two to three times larger (Section~\ref{sec:results-real}, fourth finding).

Unlike the reference, which draws every proposal from one model, EvoScore changes the model during the search (Section~\ref{sec:search}), and the logs show that this mechanism was active in every run. In the main study, the logged fitness trajectories imply 293 stagnation-triggered switches over the 72 runs, i.e., 2 to 5 per run. To quantify the use of the pool in more detail, the model answering each request was recorded in three repetitions of the search on eight of the classes (24 runs, 5,093 requests). All three models were used in every run, with 4 to 15 switches per run (7.8 on average), of which 54\% were triggered by stagnation and 46\% by malformed proposals. DeepSeek V4 Flash answered 44\% of the requests, Gemini 2.5 Flash Lite 30\%, and Qwen3 Coder 26\%, and 45\%, 31\%, and 49\% of their proposals, respectively, were registered as valid programs. More than half of the requests were thus answered by the second and the third model of the pool, i.e., the search did not reduce to its first model.

\section{Discussion}
\label{sec:discussion}

\subsection{Mechanisms of performance gains and structural interpretability}

The empirical performance gains achieved by the evolved heuristics stem primarily from tie-breaking refinement. While classical seed heuristics perform reliably on primary placement objectives, they lack a principled mechanism to distinguish among candidate positions that tie on those objectives. Evolutionary optimization bridges this gap by injecting targeted terms designed specifically to resolve such ties.

Consequently, the advantage is greatest where ties occur frequently and carry significant geometric impact—specifically in heterogeneous item mixes and long episodes (Section~\ref{sec:results-dims}). Conversely, homogeneous streams of large items exhibit the smallest gains, as a small number of large rectangles leaves little geometric freedom for tie-break optimization. This mechanism also explains why the generated algorithms remain interpretable: each component added to the scoring formula—such as wall affinity, waste penalty, support bonus, or orientation preference—corresponds to a parameter that a human decision maker would easily recognize.

\subsection{Methodological scope and relation to the literature}

Our findings align well with the rigorous studies discussed in Section \ref{LLM_driven_heuristics}. Consistent with Sim, Renau, and Hart \cite{ref34}, we do not claim to have discovered a universal heuristic; rather, each evolved heuristic is tailored to a specific instance class and tested exclusively on that class. Our protocol ensures that any overfitting to the evolutionary pool becomes apparent during validation and is not rewarded on the held-out test pool. Following Herrmann and Pallez  \cite{ref35}, we do not identify readability with human comprehension; instead, we establish—on a program-by-program basis—the modifications made relative to a known seed, which manifest as additive tie-breaking terms applied to a classical framework. This contribution therefore represents a discovery that remains valid within a leak-free protocol, rather than a disruptive, ungrounded optimization paradigm for packing problems.

\subsection{Practical deployment and operational profile}

From an engineering and deployment perspective, the proposed approach offers a highly efficient operational profile.
Heuristic discovery is a one-time offline procedure for each distribution class, whereas the deployed model is a CPU-only scoring function with microsecond-level execution time. This allows it to replace the placement rule in existing packing or cutting software without requiring specialized hardware or new infrastructure. Furthermore, the validation threshold acts as an intuitive acceptance test that can be governed by a non-specialist: an evolved rule is deployed only if it outperforms the incumbent current classical heuristic on held-out data; otherwise, the existing heuristic remains active. If the underlying item distribution shifts significantly over time, operators simply re-fit the distribution parameters and re-execute the evolutionary search.

\subsection{Data efficiency and privacy-preserving synthesis}

A significant advantage of the proposed framework is its sufficiency of distribution parameters, which effectively eliminates dependencies on raw operational data. Evolving heuristics against a parametric model fitted to a real product catalog yields performance statistically equivalent (within a 1\% margin) to evolving directly on raw item streams, while consistently outperforming all classical baselines on real held-out instances.

This characteristic enables privacy-preserving synthesis. Because summary statistics are frequently accessible even when raw order logs are proprietary, sensitive, or volatile, third-party developers and internal optimization teams can synthesize highly tailored, high-performing heuristics without ever exposing or directly accessing raw operational instances.

\subsection{Threads of validity}

Our study reveals some limitations. First of all, results depend on the execution environment and LLMs act in non-deterministic way, therefore re-running
the experiments may not reproduce the exact same results. Each instance class was evolved once with a single random seed, leaving the variance of search outcomes across multiple seeds unmeasured. The four extensions over FunSearch (classical warm starts, model switching, context-conditioned prompting, and validation floor) were introduced jointly, meaning their individual contributions are not isolated.

A notable threat to validity is that our study focuses on a 72 instance classes and instances built upon the catalog of real-world items from one industry area, which may not be representative of all possible datasets,
so results should be interpreted in the context of our dataset. Evolved heuristics were evaluated solely on held-out instances from their own training class; cross-class transfer and the performance of a single generalist heuristic evolved across all classes were not tested (in the sense of Sim, Renau, and Hart \cite{ref34}).

Finally, while the discovered programs are readable and their tie-break terms can be named, human comprehension was not empirically tested to verify whether a reader can accurately predict program behavior on unseen instances. 
\section{Conclusions}
\label{sec:conclusions}

In this paper, we demonstrated that LLM-guided evolutionary program search, utilizing a scoring-function representation over a canonical candidate set, yields superior online 2D strip-packing heuristics. Building upon FunSearch, we extended the framework with four methodological enhancements: classical warm starts, dynamic model switching, context-conditioned prompting, and a validation floor. We benchmarked the evolved heuristics against eleven classical baselines and standard FunSearch across both synthetic and real-world logistics datasets. Our results confirm the effectiveness of EvoScore: the evolved heuristics outperformed all eleven online baselines in paired comparisons across 7,200 synthetic test episodes, winning 60 out of 72 instance classes against the strongest class-specific competitor. On real-world logistics data, EvoScore not only achieved the highest mean utilization, but also demonstrated that evolving heuristics on fitted distribution parameters yields statistically equivalent performance to evolving directly on real item streams. Notably, across diverse problem classes, the evolutionary process exhibits a consistent structural pattern—preserving the core decision rule of the classical seed while synthesizing compact, interpretable tie-breaking terms.

Future work can proceed along three primary directions: methodology, evaluation, and problem domains. From a methodological point of view, it could be interesting to conduct an ablation study to isolate the contributions of the four extensions alongside multi-seed search replications, as well as seeding the search with syntactically distinct implementations of the same classical rule to evaluate representation sensitivity. Incorporating online fine-tuning of the LLM generator during the evolutionary loop may further improve search efficiency and solution quality. Additionally, integrating out-of-distribution detection with a classical fallback mechanism could enable special handling of such instances using a portfolio of heuristics, while adopting risk-sensitive fitness functions (e.g., penalizing variance or tail risk alongside mean performance) can mitigate severe worst-case failures. Regarding evaluation, future experiments shall expand to additional data sets of synthetic instances as well as real-world logistics data, exhibiting distinct structural properties such as clustered order profiles. Finally, within the broader problem domain, the decoupled architecture of EvoScore allows for straightforward extension to related cutting and packing problems—including 2D bin packing, pallet loading, and container loading—by replacing the candidate generator while leaving the evolutionary search process unchanged.
\bibliographystyle{apalike}
\bibliography{biblio}

@article{ref1,
  author  = {Dyckhoff, Harald},
  title   = {A typology of cutting and packing problems},
  journal = {European Journal of Operational Research},
  year    = {1990},
  volume  = {44},
  number  = {2},
  pages   = {145--159},
  doi     = {10.1016/0377-2217(90)90350-K}
}

@article{ref2,
  author  = {W{\"a}scher, Gerhard and Hau{\ss}ner, Heike and Schumann, Holger},
  title   = {An improved typology of cutting and packing problems},
  journal = {European Journal of Operational Research},
  year    = {2007},
  volume  = {183},
  number  = {3},
  pages   = {1109--1130},
  doi     = {10.1016/j.ejor.2005.12.047}
}

@article{ref3,
  author  = {Neuenfeldt J{\'u}nior, Alvaro and Silva, Elsa and Francescatto, Matheus and Rosa, Carmen Brum and Siluk, Julio},
  title   = {The rectangular two-dimensional strip packing problem real-life practical constraints: A bibliometric overview},
  journal = {Computers \& Operations Research},
  year    = {2022},
  volume  = {137},
  pages   = {105521},
  doi     = {10.1016/j.cor.2021.105521}
}

@article{ref4,
  author  = {Crainic, Teodor Gabriel and Perboli, Guido and Tadei, Roberto},
  title   = {Extreme point-based heuristics for three-dimensional bin packing},
  journal = {{INFORMS} Journal on Computing},
  year    = {2008},
  volume  = {20},
  number  = {3},
  pages   = {368--384},
  doi     = {10.1287/ijoc.1070.0250}
}

@article{ref5,
  author  = {Baker, Brenda S. and Coffman, Jr., E. G. and Rivest, Ronald L.},
  title   = {Orthogonal packings in two dimensions},
  journal = {{SIAM} Journal on Computing},
  year    = {1980},
  volume  = {9},
  number  = {4},
  pages   = {846--855},
  doi     = {10.1137/0209064}
}

@article{ref6,
  author  = {Burke, Edmund K. and Kendall, Graham and Whitwell, Glenn},
  title   = {A new placement heuristic for the orthogonal stock-cutting problem},
  journal = {Operations Research},
  year    = {2004},
  volume  = {52},
  number  = {4},
  pages   = {655--671},
  doi     = {10.1287/opre.1040.0109}
}

@techreport{ref7,
  author = {Jyl{\"a}nki, Jukka},
  title  = {A thousand ways to pack the bin -- A practical approach to two-dimensional rectangle bin packing},
  type   = {Technical report},
  year   = {2010},
  month  = feb,
  url    = {https://github.com/juj/RectangleBinPack/blob/master/RectangleBinPack.pdf},
  note   = {Dated 27 February 2010; distributed with the RectangleBinPack repository}
}

@article{ref8,
  author  = {Lodi, Andrea and Martello, Silvano and Vigo, Daniele},
  title   = {Heuristic and metaheuristic approaches for a class of two-dimensional bin packing problems},
  journal = {{INFORMS} Journal on Computing},
  year    = {1999},
  volume  = {11},
  number  = {4},
  pages   = {345--357},
  doi     = {10.1287/ijoc.11.4.345}
}

@article{ref9,
  author  = {Wolpert, David H. and Macready, William G.},
  title   = {No free lunch theorems for optimization},
  journal = {{IEEE} Transactions on Evolutionary Computation},
  year    = {1997},
  volume  = {1},
  number  = {1},
  pages   = {67--82},
  doi     = {10.1109/4235.585893}
}

@article{ref10,
  author  = {Romera-Paredes, Bernardino and Barekatain, Mohammadamin and Novikov, Alexander and Balog, Matej and Kumar, M. Pawan and Dupont, Emilien and Ruiz, Francisco J. R. and Ellenberg, Jordan S. and Wang, Pengming and Fawzi, Omar and Kohli, Pushmeet and Fawzi, Alhussein},
  title   = {Mathematical discoveries from program search with large language models},
  journal = {Nature},
  year    = {2024},
  volume  = {625},
  number  = {7995},
  pages   = {468--475},
  doi     = {10.1038/s41586-023-06924-6}
}

@article{ref11,
  author  = {Kaleta, Mariusz and Ko{\l}odziejczyk, Waldemar and Zoltowska, Izabela},
  title   = {Transformer-based placement heuristic for online {2D} strip packing problem},
  journal = {Computers \& Industrial Engineering},
  year    = {2025},
  volume  = {210},
  pages   = {111464},
  doi     = {10.1016/j.cie.2025.111464}
}

@article{ref14,
  author  = {Baker, Brenda S. and Schwarz, Jerald S.},
  title   = {Shelf algorithms for two-dimensional packing problems},
  journal = {{SIAM} Journal on Computing},
  year    = {1983},
  volume  = {12},
  number  = {3},
  pages   = {508--525},
  doi     = {10.1137/0212033}
}

@article{ref15,
  author  = {Csirik, J{\'a}nos and Woeginger, Gerhard J.},
  title   = {Shelf algorithms for on-line strip packing},
  journal = {Information Processing Letters},
  year    = {1997},
  volume  = {63},
  number  = {4},
  pages   = {171--175},
  doi     = {10.1016/S0020-0190(97)00120-8}
}

@article{ref17,
  author  = {Ye, Deshi and Han, Xin and Zhang, Guochuan},
  title   = {A note on online strip packing},
  journal = {Journal of Combinatorial Optimization},
  year    = {2009},
  volume  = {17},
  number  = {4},
  pages   = {417--423},
  doi     = {10.1007/s10878-007-9125-x}
}

@article{ref21,
  author  = {Ali, Sara and Ramos, Ant{\'o}nio Galr{\~a}o and Carravilla, Maria Ant{\'o}nia and Oliveira, Jos{\'e} Fernando},
  title   = {Heuristics for online three-dimensional packing problems and algorithm selection framework for semi-online with full look-ahead},
  journal = {Applied Soft Computing},
  year    = {2024},
  volume  = {151},
  pages   = {111168},
  doi     = {10.1016/j.asoc.2023.111168}
}

@article{ref22,
  author  = {Branke, J{\"u}rgen and Nguyen, Su and Pickardt, Christoph W. and Zhang, Mengjie},
  title   = {Automated design of production scheduling heuristics: A review},
  journal = {{IEEE} Transactions on Evolutionary Computation},
  year    = {2016},
  volume  = {20},
  number  = {1},
  pages   = {110--124},
  doi     = {10.1109/TEVC.2015.2429314}
}

@inproceedings{ref25,
  author    = {Liu, Fei and Tong, Xialiang and Yuan, Mingxuan and Lin, Xi and Luo, Fu and Wang, Zhenkun and Lu, Zhichao and Zhang, Qingfu},
  title     = {Evolution of heuristics: Towards efficient automatic algorithm design using large language model},
  booktitle = {Proceedings of the 41st International Conference on Machine Learning ({ICML} 2024)},
  editor    = {Salakhutdinov, Ruslan and Kolter, Zico and Heller, Katherine and Weller, Adrian and Oliver, Nuria and Scarlett, Jonathan and Berkenkamp, Felix},
  series    = {Proceedings of Machine Learning Research},
  volume    = {235},
  pages     = {32201--32223},
  publisher = {PMLR},
  year      = {2024},
  url       = {https://proceedings.mlr.press/v235/liu24bs.html}
}

@inproceedings{ref26,
  author    = {Ye, Haoran and Wang, Jiarui and Cao, Zhiguang and Berto, Federico and Hua, Chuanbo and Kim, Haeyeon and Park, Jinkyoo and Song, Guojie},
  title     = {{ReEvo}: Large language models as hyper-heuristics with reflective evolution},
  booktitle = {Advances in Neural Information Processing Systems 37 ({NeurIPS} 2024)},
  editor    = {Globerson, A. and Mackey, L. and Belgrave, D. and Fan, A. and Paquet, U. and Tomczak, J. and Zhang, C.},
  volume    = {37},
  pages     = {43571--43608},
  publisher = {Curran Associates, Inc.},
  year      = {2024},
  doi       = {10.52202/079017-1381},
  url       = {https://proceedings.neurips.cc/paper_files/paper/2024/hash/4ced59d480e07d290b6f29fc8798f195-Abstract-Conference.html}
}

@inproceedings{ref27,
  author    = {{\v{S}}urina, Anja and Mansouri, Amin and Quaedvlieg, Lars C. P. M. and Seddas, Amal and Viazovska, Maryna and Abbe, Emmanuel and Gulcehre, Caglar},
  title     = {Algorithm discovery with {LLMs}: Evolutionary search meets reinforcement learning},
  booktitle = {Second Conference on Language Modeling ({COLM} 2025)},
  year      = {2025},
  url       = {https://openreview.net/forum?id=dNW3RGW0gi},
  eprint    = {2504.05108},
  archivePrefix = {arXiv},
  primaryClass  = {cs.AI}
}

@article{ref28,
  author  = {Xu, Jia-Wei and Wei, Feng-Feng and Chen, Wei-Neng},
  title   = {{RoCo}: Role-based {LLMs} collaboration for automatic heuristic design},
  journal = {{IEEE} Transactions on Artificial Intelligence},
  year    = {2026},
  pages   = {1--15},
  doi     = {10.1109/TAI.2026.3730656},
  note    = {Early access}
}

@misc{ref29,
  author = {Lv, Haoze and Lu, Ning and Zhou, Ziang and Ong, Yew-Soon and Liu, Shengcai},
  title  = {{AHD} Agent: Agentic reinforcement learning for automatic heuristic design},
  year   = {2026},
  eprint = {2605.08756},
  archivePrefix = {arXiv},
  primaryClass  = {cs.AI},
  url    = {https://arxiv.org/abs/2605.08756},
  note   = {arXiv preprint}
}

@misc{ref30,
  author = {Yang, Xianliang and Zhang, Ling and Qian, Haolong and Song, Lei and Bian, Jiang},
  title  = {{HeurAgenix}: Leveraging {LLMs} for solving complex combinatorial optimization challenges},
  year   = {2025},
  eprint = {2506.15196},
  archivePrefix = {arXiv},
  primaryClass  = {cs.AI},
  url    = {https://arxiv.org/abs/2506.15196},
  note   = {arXiv preprint}
}

@misc{ref31,
  author = {Zhou, XinLei and Huang, Jin and Yang, Jie and Li, Xinyu and Gao, Liang},
  title  = {{DSevolve}: Enabling real-time adaptive scheduling on dynamic flexible job shop with {LLM}-evolved heuristic portfolios},
  year   = {2026},
  eprint = {2603.27628},
  archivePrefix = {arXiv},
  primaryClass  = {cs.AI},
  url    = {https://arxiv.org/abs/2603.27628},
  note   = {arXiv preprint}
}

@misc{ref32,
  author = {Qiu, Junhao and Zhuang, Haoyang and Liu, Fei and Liu, Jianjun and Zhang, Qingfu},
  title  = {{EvoDR}: Evolving dispatching rules via large language model for dynamic flexible assembly flow shop scheduling},
  year   = {2026},
  eprint = {2601.15738},
  archivePrefix = {arXiv},
  primaryClass  = {cs.NE},
  url    = {https://arxiv.org/abs/2601.15738},
  note   = {arXiv preprint}
}

@article{ref33,
  author  = {{\c{C}}etinkaya, {\.I}brahim O{\u{g}}uz and B{\"u}y{\"u}ktahtak{\i}n, {\.I}. Esra and Shojaee, Parshin and Reddy, Chandan K.},
  title   = {Discovering heuristics with large language models ({LLMs}) for mixed-integer programs: Single-machine scheduling},
  journal = {Computers \& Operations Research},
  year    = {2026},
  volume  = {186},
  pages   = {107325},
  doi     = {10.1016/j.cor.2025.107325}
}

@inproceedings{ref34,
  author    = {Sim, Kevin and Renau, Quentin and Hart, Emma},
  title     = {Beyond the hype: Benchmarking {LLM}-evolved heuristics for bin packing},
  booktitle = {Applications of Evolutionary Computation ({EvoApplications} 2025), Part {II}},
  editor    = {Garc{\'i}a-S{\'a}nchez, Pablo and Hart, Emma and Thomson, Sarah L.},
  series    = {Lecture Notes in Computer Science},
  volume    = {15613},
  pages     = {386--402},
  publisher = {Springer},
  address   = {Cham},
  year      = {2025},
  doi       = {10.1007/978-3-031-90065-5_24}
}

@article{ref35,
  author  = {Herrmann, Julien and Pallez, Guillaume},
  title   = {An in-depth study of {LLM} contributions to the bin packing problem},
  journal = {{ACM} Transactions on Evolutionary Learning and Optimization},
  year    = {2026},
  doi     = {10.1145/3821574},
  note    = {Article 3821574, published online 18 June 2026}
}

@inproceedings{ref36,
  author    = {Huq, Syed Mahbubul and Brito, Daniel and Sikar, Daniel and Child, Chris and Weyde, Tillman and Mojumder, Rajesh},
  title     = {Evaluating {LLMs} for combinatorial optimization: One-phase and two-phase heuristics for {2D} bin-packing},
  booktitle = {{NeurIPS} 2025 Workshop on Evaluating the Evolving {LLM} Lifecycle: Benchmarks, Emergent Abilities, and Scaling},
  year      = {2025},
  url       = {https://openreview.net/forum?id=Mz5uv4orm8},
  eprint    = {2509.22255},
  archivePrefix = {arXiv},
  primaryClass  = {cs.AI}
}

@misc{ref37,
  author = {Quan, Guorui and Sun, Mingfei and L{\'o}pez-Ib{\'a}{\~n}ez, Manuel},
  title  = {Re-evaluating {LLM}-based heuristic search: A case study on the {3D} packing problem},
  year   = {2025},
  eprint = {2509.02297},
  archivePrefix = {arXiv},
  primaryClass  = {cs.AI},
  url    = {https://arxiv.org/abs/2509.02297},
  note   = {arXiv preprint}
}

@article{ref38,
  author  = {Kaleta, Mariusz and {\'S}liwi{\'n}ski, Tomasz},
  title   = {Neural-driven heuristic for strip packing trained with black-box optimization},
  journal = {International Journal of Electronics and Telecommunications},
  year    = {2025},
  volume  = {71},
  number  = {2},
  pages   = {387--396},
  doi     = {10.24425/ijet.2025.153584}
}

@article{ref39,
  author  = {Kaleta, Mariusz and {\'S}liwi{\'n}ski, Tomasz},
  title   = {Improving neural-based heuristics for {2D} strip packing through local search in heuristic space},
  journal = {Neural Computing and Applications},
  year    = {2026},
  volume  = {38},
  number  = {9},
  pages   = {337},
  doi     = {10.1007/s00521-026-12013-2}
}

@article{ref40,
  author  = {Xu, Yang and Yang, Zhouwang},
  title   = {{GraphPack}: A reinforcement learning algorithm for strip packing problem using graph neural network},
  journal = {Journal of Circuits, Systems and Computers},
  year    = {2024},
  volume  = {33},
  number  = {8},
  pages   = {2450139},
  doi     = {10.1142/S0218126624501391}
}

@misc{ref41,
  author = {Huang, Qi and Ye, Furong and Shahane, Ananta and B{\"a}ck, Thomas and van Stein, Niki},
  title  = {From heuristic selection to automated algorithm design: {LLMs} benefit from strong priors},
  year   = {2026},
  eprint = {2603.02792},
  archivePrefix = {arXiv},
  primaryClass  = {cs.LG},
  url    = {https://arxiv.org/abs/2603.02792},
  note   = {arXiv preprint}
}

@inproceedings{ref42,
  author    = {Wu, Xuan and Wang, Di and Wu, Chunguo and Wen, Lijie and Miao, Chunyan and Xiao, Yubin and Zhou, You},
  title     = {Efficient heuristics generation for solving combinatorial optimization problems using large language models},
  booktitle = {Proceedings of the 31st {ACM} {SIGKDD} Conference on Knowledge Discovery and Data Mining V.2 ({KDD} '25)},
  pages     = {3228--3239},
  publisher = {ACM},
  year      = {2025},
  doi       = {10.1145/3711896.3736923}
}

@inproceedings{ref43,
  author    = {Liu, Fei and Liu, Yilu and Zhang, Qingfu and Tong, Xialiang and Yuan, Mingxuan},
  title     = {{EoH-S}: Evolution of heuristic set using {LLMs} for automated heuristic design},
  booktitle = {Proceedings of the {AAAI} Conference on Artificial Intelligence},
  volume    = {40},
  number    = {43},
  pages     = {37090--37098},
  year      = {2026},
  doi       = {10.1609/aaai.v40i43.41038},
  url       = {https://ojs.aaai.org/index.php/AAAI/article/view/41038}
}

@misc{ref44,
  author = {Karimi, Pantea and Rouhana, Dany and Namyar, Pooria and Kakarla, Siva Kesava Reddy and Arun, Venkat and Arzani, Behnaz},
  title  = {Robust heuristic algorithm design with {LLMs}},
  year   = {2025},
  eprint = {2510.08755},
  archivePrefix = {arXiv},
  primaryClass  = {cs.AI},
  url    = {https://arxiv.org/abs/2510.08755},
  note   = {arXiv preprint; also Microsoft Research technical report MSR-TR-2025-44}
}

@misc{ref45,
  author = {Novikov, Alexander and V{\~u}, Ng{\^a}n and Eisenberger, Marvin and Dupont, Emilien and Huang, Po-Sen and Wagner, Adam Zsolt and Shirobokov, Sergey and Kozlovskii, Borislav and Ruiz, Francisco J. R. and Mehrabian, Abbas and Kumar, M. Pawan and See, Abigail and Chaudhuri, Swarat and Holland, George and Davies, Alex and Nowozin, Sebastian and Kohli, Pushmeet and Balog, Matej},
  title  = {{AlphaEvolve}: A coding agent for scientific and algorithmic discovery},
  year   = {2025},
  eprint = {2506.13131},
  archivePrefix = {arXiv},
  primaryClass  = {cs.AI},
  url    = {https://arxiv.org/abs/2506.13131},
  note   = {arXiv preprint}
}

@inproceedings{ref46,
  author    = {Lange, Robert Tjarko and Imajuku, Yuki and Cetin, Edoardo},
  title     = {{ShinkaEvolve}: Towards open-ended and sample-efficient program evolution},
  booktitle = {The Fourteenth International Conference on Learning Representations ({ICLR} 2026)},
  year      = {2026},
  url       = {https://openreview.net/forum?id=lKEdGCoDNC},
  eprint    = {2509.19349},
  archivePrefix = {arXiv},
  primaryClass  = {cs.CL}
}

@misc{ref47,
  author = {Zhang, Shaofeng and Liu, Shengcai and Lu, Ning and Wu, Jiahao and Liu, Ji and Ong, Yew-Soon and Tang, Ke},
  title  = {{LLM}-driven instance-specific heuristic generation and selection},
  year   = {2025},
  eprint = {2506.00490},
  archivePrefix = {arXiv},
  primaryClass  = {cs.NE},
  url    = {https://arxiv.org/abs/2506.00490},
  hidenote   = {arXiv preprint; no IEEE TETCI record found as of October 2026}
}

@article{ref48,
  author  = {Hottung, Andr{\'e} and Berto, Federico and Hua, Chuanbo and Gast Zepeda, Nayeli and Wetzel, Daniel and R{\"o}mer, Michael and Ye, Haoran and Zago, Davide and Poli, Michael and Massaroli, Stefano and Park, Jinkyoo and Tierney, Kevin},
  title   = {{VRPAgent}: {LLM}-driven discovery of heuristic operators for vehicle routing problems},
  journal = {Transactions on Machine Learning Research},
  year    = {2026},
  issn    = {2835-8856},
  url     = {https://openreview.net/forum?id=2cYbzycSbz},
  eprint  = {2510.07073},
  archivePrefix = {arXiv},
  primaryClass  = {cs.AI}
}

@misc{ref49,
  author = {Michailidis, Kostis and Tsouros, Dimos and Dang, Nguyen and Guns, Tias},
  title  = {{LLM}-guided evolutionary search for constraint model reformulation to improve solver efficiency},
  year   = {2026},
  eprint = {2607.28268},
  archivePrefix = {arXiv},
  primaryClass  = {cs.AI},
  url    = {https://arxiv.org/abs/2607.28268},
  note   = {arXiv preprint}
}

@article{ref50,
  author  = {Weindel, Franziska and Heckel, Reinhard},
  title   = {{LLM}-guided search for deletion-correcting codes},
  journal = {Transactions on Machine Learning Research},
  year    = {2026},
  issn    = {2835-8856},
  url     = {https://openreview.net/forum?id=qZ69Ozpo6v},
  eprint  = {2504.00613},
  archivePrefix = {arXiv},
  primaryClass  = {cs.AI}
}

@article{ref51,
  author  = {von Hippel, Matt and Wilhelm, Matthias},
  title   = {Refining integration-by-parts reduction of {Feynman} integrals with machine learning},
  journal = {Journal of High Energy Physics},
  year    = {2025},
  volume  = {2025},
  number  = {5},
  pages   = {185},
  doi     = {10.1007/JHEP05(2025)185}
}

@inproceedings{ref52,
  author    = {Lv, Junjin and Cui, Chenggang and Zhang, Shaodi and Chen, Hui and Gong, Chunyang and Liu, Jiaming},
  title     = {Automated heuristic design for unit commitment using large language models},
  booktitle = {2025 International Power and Electrical Engineering Conference ({IPEE})},
  pages     = {1--4},
  publisher = {IEEE},
  year      = {2025},
  doi       = {10.1109/IPEE64915.2025.11452800}
}

@misc{ref53,
  author = {Liu, Fei and Zhang, Rui and Xie, Zhuoliang and Sun, Rui and Li, Kai and Hu, Qinglong and Guo, Ping and Lin, Xi and Tong, Xialiang and Yuan, Mingxuan and Wang, Zhenkun and Lu, Zhichao and Zhang, Qingfu},
  title  = {{LLM4AD}: A platform for algorithm design with large language model},
  year   = {2024},
  eprint = {2412.17287},
  archivePrefix = {arXiv},
  primaryClass  = {cs.AI},
  url    = {https://arxiv.org/abs/2412.17287},
  note   = {arXiv preprint}
}

@article{ref54,
  author  = {Martello, Silvano and Pisinger, David and Vigo, Daniele},
  title   = {The three-dimensional bin packing problem},
  journal = {Operations Research},
  year    = {2000},
  volume  = {48},
  number  = {2},
  pages   = {256--267},
  doi     = {10.1287/opre.48.2.256.12386}
}

@article{hifi2026critical,
  author  = {Hifi, Mhand and Zhao, Juntao},
  title   = {A critical review on cutting and packing problems: Recent advancements, challenges and applications},
  journal = {Journal of the Operational Research Society},
  year    = {2026},
  pages   = {1--47},
  doi     = {10.1080/01605682.2026.2717320},
  note    = {Published online 31 August 2026}
}

@article{christensen2017approximation,
  author  = {Christensen, Henrik I. and Khan, Arindam and Pokutta, Sebastian and Tetali, Prasad},
  title   = {Approximation and online algorithms for multidimensional bin packing: A survey},
  journal = {Computer Science Review},
  year    = {2017},
  volume  = {24},
  pages   = {63--79},
  doi     = {10.1016/j.cosrev.2016.12.001}
}

@article{wei_open_2019,
  author  = {Wei, Lijun and Wang, Yongsheng and Cheng, Huibing and Huang, Jian},
  title   = {An open space based heuristic for the {2D} strip packing problem with unloading constraints},
  journal = {Applied Mathematical Modelling},
  year    = {2019},
  volume  = {70},
  pages   = {67--81},
  doi     = {10.1016/j.apm.2019.01.022}
}

@article{Boyar2017,
  author  = {Boyar, Joan and Favrholdt, Lene M. and Kudahl, Christian and Larsen, Kim S. and Mikkelsen, Jesper W.},
  title   = {Online algorithms with advice: A survey},
  journal = {{ACM} Computing Surveys},
  year    = {2017},
  volume  = {50},
  number  = {2},
  pages   = {1--34},
  doi     = {10.1145/3056461}
}

@article{oliveira2016survey,
  author  = {Oliveira, Jos{\'e} Fernando and Neuenfeldt J{\'u}nior, Alvaro and Silva, Elsa and Carravilla, Maria Ant{\'o}nia},
  title   = {A survey on heuristics for the two-dimensional rectangular strip packing problem},
  journal = {Pesquisa Operacional},
  year    = {2016},
  volume  = {36},
  number  = {2},
  pages   = {197--226},
  doi     = {10.1590/0101-7438.2016.036.02.0197}
}

@article{Mezghani2023,
  author  = {Mezghani, Salma and Haddar, Boukthir and Chabchoub, Habib},
  title   = {The evolution of rectangular bin packing problem -- A review of research topics, applications, and cited papers},
  journal = {Journal of Industrial and Management Optimization},
  year    = {2023},
  volume  = {19},
  number  = {5},
  pages   = {3329--3361},
  doi     = {10.3934/jimo.2022088}
}

@article{asta2016champ,
  author  = {Asta, Shahriar and {\"O}zcan, Ender and Parkes, Andrew J.},
  title   = {{CHAMP}: Creating heuristics via many parameters for online bin packing},
  journal = {Expert Systems with Applications},
  year    = {2016},
  volume  = {63},
  pages   = {208--221},
  doi     = {10.1016/j.eswa.2016.07.005}
}

@article{liu2026systematic,
  author  = {Liu, Fei and Yao, Yiming and Guo, Ping and Yang, Zhiyuan and Lin, Xi and Zhao, Zhe and Tong, Xialiang and Mao, Kun and Lu, Zhichao and Wang, Zhenkun and Yuan, Mingxuan and Zhang, Qingfu},
  title   = {A systematic survey on large language models for algorithm design},
  journal = {{ACM} Computing Surveys},
  year    = {2026},
  volume  = {58},
  number  = {8},
  pages   = {1--32},
  doi     = {10.1145/3787585}
}

@article{yang2025learning,
  author  = {Yang, Shuo and Song, Shuai and Song, Ran and Cheng, Jiyu and Li, Yibin and Zhang, Wei},
  title   = {Learning packing-and-unpacking synergistic policy via {LLM}-guided {DRL} for robust online robotic packing},
  journal = {Advanced Engineering Informatics},
  year    = {2025},
  volume  = {68},
  pages   = {103572},
  doi     = {10.1016/j.aei.2025.103572}
}

@article{wong2024integrating,
  title={Integrating heuristic methods with deep reinforcement learning for online 3D bin-packing optimization},
  author={Wong, Ching-Chang and Tsai, Tai-Ting and Ou, Can-Kun},
  journal={Sensors},
  volume={24},
  number={16},
  pages={5370},
  year={2024},
  doi={10.3390/s24165370}
}

@ARTICLE{xiong,
  author={Xiong, Heng and Guo, Changrong and Peng, Jian and Ding, Kai and Chen, Wenjie and Qiu, Xuchong and Bai, Long and Xu, Jianfeng},
  journal={IEEE Robotics and Automation Letters}, 
  title={{GOPT}: Generalizable Online 3D Bin Packing via Transformer-Based Deep Reinforcement Learning}, 
  year={2024},
  volume={9},
  number={11},
  pages={10335-10342},
  doi={10.1109/LRA.2024.3468161}}

@misc{deepseekv4flash,
  author       = {{DeepSeek}},
  title        = {{DeepSeek-V4-Flash}: Model card},
  year         = {2026},
  howpublished = {\url{https://huggingface.co/deepseek-ai/DeepSeek-V4-Flash}},
  note         = {Accessed: 19 November 2026}
}

@misc{qwen3coder,
  author       = {{Qwen Team}},
  title        = {{Qwen3-Coder}: Agentic coding in the world},
  year         = {2025},
  howpublished = {\url{https://qwenlm.github.io/blog/qwen3-coder/}},
  note         = {Accessed: 19 November 2026}
}

@misc{gemini25flashlite,
  author       = {{Google}},
  title        = {{Gemini 2.5 Flash-Lite}: Model card},
  year         = {2025},
  howpublished = {\url{https://storage.googleapis.com/deepmind-media/Model-Cards/Gemini-2-5-Flash-Lite-Model-Card.pdf}},
  note         = {Accessed: 19 November 2026}
}
\onecolumn
\lstset{basicstyle=\ttfamily\footnotesize}
\appendix
\section{Evolved programs of the four examples}
\label{app:programs}

Listings~\ref{lst:ex-a}--\ref{lst:ex-d} reproduce the four evolved programs discussed in Section~\ref{sec:results-seeds} as they were returned by the search and selected under the validation floor of Section~\ref{sec:search}. Only the docstring of the function header, which is identical in all programs and repeats the interface of Section~\ref{sec:candidates}, has been removed; the code is otherwise unchanged, including the inactive variable \texttt{gap\_penalty} in Listing~\ref{lst:ex-c}. Every program has the signature of the seeds: it receives the strip width, the placed items, one candidate placement and the full candidate set, and returns a score, and the candidate with the highest score is placed.

\begin{lstlisting}[caption={Example 1, Lowest-Top lineage, class moderate aspect ratio, medium width, high aspect-ratio variance, high width variance, 10 items.},label={lst:ex-a}]
def score_placement(strip_width, placed_items, candidate, all_candidates):
  top = candidate["y"] + candidate["h"]
  min_top = min(c["y"] + c["h"] for c in all_candidates)
  max_top = max(c["y"] + c["h"] for c in all_candidates)
  span = max_top - min_top + 1
  width = candidate["w"]
  left = candidate["x"]
  right = left + width
  left_space = left
  right_space = strip_width - right
  side_penalty = -min(left_space, right_space) * 0.1
  height_penalty = -(top - min_top) / span * 10000
  x_penalty = -candidate["x"] * 0.01
  return height_penalty + side_penalty + x_penalty
\end{lstlisting}

\begin{lstlisting}[caption={Example 2, Best-Fit lineage, class square, large, high aspect-ratio variance, low width variance, 10 items.},label={lst:ex-b}]
def score_placement(strip_width, placed_items, candidate, all_candidates):
  current_max = max((p["y"] + p["h"] for p in placed_items), default=0)
  height_increase = max(0, candidate["y"] + candidate["h"] - current_max)

  # Calculate how well the item sits in a "corner" - items that touch both bottom and side
  touches_left_wall = candidate["x"] == 0
  touches_right_wall = candidate["x"] + candidate["w"] == strip_width
  touches_bottom = candidate["y"] == 0
  touches_placed_item_below = any(
    p["x"] <= candidate["x"] + candidate["w"] and 
    p["x"] + p["w"] >= candidate["x"] and
    p["y"] + p["h"] == candidate["y"]
    for p in placed_items
  )

  corner_score = 0
  if (touches_left_wall or touches_right_wall) and (touches_bottom or touches_placed_item_below):
    corner_score = 1000

  # Minimize height increase, then prefer lower placements, then leftmost
  return -(height_increase * 10000 + candidate["y"] * 100 + candidate["x"] - corner_score)
\end{lstlisting}

\begin{lstlisting}[caption={Example 3, Best-Fit lineage, class square, large, low aspect-ratio variance, high width variance, 100 items.},label={lst:ex-c}]
def score_placement(strip_width, placed_items, candidate, all_candidates):
  # More compact variant - reuse v0's logic but adjust weights for balance
  w, h = candidate["w"], candidate["h"]
  x, y = candidate["x"], candidate["y"]

  current_max = max((p["y"] + p["h"] for p in placed_items), default=0)
  height_increase = max(0, y + h - current_max)

  right_margin = strip_width - (x + w)

  # Gap penalty based on items above candidate
  gap_penalty = 0.0
  for p in placed_items:
    if p["x"] < x + w and p["x"] + p["w"] > x and p["y"] + p["h"] <= y:
      gap_overlap = min(p["x"] + p["w"], x + w) - max(p["x"], x)
      if gap_overlap > 0:
        gap_penalty += gap_overlap

  # Edge bonus for touching left/right or being close to already placed items
  edge_bonus = 0.0
  if x == 0 or x + w == strip_width:
    edge_bonus = 50.0

  # Reward horizontal clustering: small gaps to neighbors
  x_gap = strip_width
  for p in placed_items:
    if p["y"] + p["h"] <= y or p["y"] >= y + h:
      # Same vertical band check
      if p["y"] < y + h and p["y"] + p["h"] > y:
        if p["x"] >= x + w:
          x_gap = min(x_gap, p["x"] - (x + w))
        elif p["x"] + p["w"] <= x:
          x_gap = min(x_gap, x - (p["x"] + p["w"]))

  cluster_bonus = -x_gap * 2 if x_gap < 20 else 0

  score = -height_increase * 10000 - y * 50 - x * 5 - right_margin * 3 
  score += edge_bonus * 10 + cluster_bonus
  return score
\end{lstlisting}

\begin{lstlisting}[caption={Example 4, Max-Contact lineage, class moderate aspect ratio, medium width, high aspect-ratio variance, low width variance, 10 items.},label={lst:ex-d}]
def score_placement(strip_width, placed_items, candidate, all_candidates):
  cx, cy, cw, ch = candidate["x"], candidate["y"], candidate["w"], candidate["h"]
  contact = 0.0
  if cy == 0:
    contact += cw
  if cx == 0:
    contact += ch
  if cx + cw == strip_width:
    contact += ch
  for p in placed_items:
    px, py, pw, ph = p["x"], p["y"], p["w"], p["h"]
    h_ov = max(0, min(cx + cw, px + pw) - max(cx, px))
    if h_ov > 0:
      if cy == py + ph:
        contact += h_ov
      if cy + ch == py:
        contact += h_ov
    v_ov = max(0, min(cy + ch, py + ph) - max(cy, py))
    if v_ov > 0:
      if cx + cw == px:
        contact += v_ov
      if cx == px + pw:
        contact += v_ov

  current_height = max([p["y"] + p["h"] for p in placed_items], default=0)
  new_height = max(current_height, cy + ch)

  if not placed_items:
    delta_height = new_height
  else:
    delta_height = new_height - min(p["y"] for p in placed_items)

  score = contact - 5.0 * delta_height

  # Penalize floating items (items with no contact at the bottom)
  if cy > 0:
      has_bottom_contact = False
      for p in placed_items:
          if p["x"] < cx + cw and p["x"] + p["w"] > cx and cy == p["y"] + p["h"]:
              has_bottom_contact = True
              break
      if not has_bottom_contact:
          score -= 50.0

  return score
\end{lstlisting}

\begin{lstlisting}[caption={Fitted $n = 100$ program of the real-data study, evolved on synthetic instances from the distribution fitted to the catalogue.},label={lst:ex-e}]
def score_placement(strip_width, placed_items, candidate, all_candidates):
  current_max = max((p["y"] + p["h"] for p in placed_items), default=0)
  height_increase = max(0, candidate["y"] + candidate["h"] - current_max)

  # Primary objective: minimize the increase in strip height.
  primary_score = -height_increase * 100000

  # Secondary objectives:
  # 1. Minimize the y-coordinate of the candidate to fill lower spaces first.
  secondary_score_y = -candidate["y"] * 10

  # 2. Minimize the x-coordinate of the candidate to encourage left filling.
  secondary_score_x = -candidate["x"] * 1

  # Tertiary objective: penalize placements that create large gaps.
  # Compute the horizontal gap to the nearest item on the left and right.
  gap_left = candidate["x"]
  gap_right = strip_width - (candidate["x"] + candidate["w"])
  min_side_gap = min(gap_left, gap_right)
  tertiary_score_gap = -min_side_gap * 5

  # Quaternary objective: aspect ratio consideration.
  aspect_ratio = candidate["h"] / candidate["w"]
  aspect_ratio_deviation = abs(aspect_ratio - 0.71)
  quaternary_score_ar = -aspect_ratio_deviation * 100

  # Combine scores with weighted priorities.
  total_score = (
      primary_score +
      secondary_score_y +
      secondary_score_x +
      tertiary_score_gap +
      quaternary_score_ar
  )

  return total_score
\end{lstlisting}

\begin{lstlisting}[caption={Direct $n = 100$ program of the real-data study, evolved on the real items.},label={lst:ex-f}]
def score_placement(strip_width, placed_items, candidate, all_candidates):
  top = candidate["y"] + candidate["h"]
  right = candidate["x"] + candidate["w"]

  # Calculate how much this placement would fill gaps in the skyline
  skyline = {}
  for item in placed_items:
    y_top = item["y"] + item["h"]
    for x in range(item["x"], item["x"] + item["w"]):
      if x not in skyline or skyline[x] < y_top:
        skyline[x] = y_top

  # Count supported perimeter points
  supported_perimeter = 0
  for x in range(candidate["x"], candidate["x"] + candidate["w"]):
    if x in skyline and skyline[x] == candidate["y"]:
      supported_perimeter += 1

  # Prefer placements that are:
  # 1. Lower in height (minimize max height)
  # 2. Further left (better packing towards left)
  # 3. More supported by existing items
  height_penalty = top * 1000
  left_bonus = -candidate["x"] * 10
  support_bonus = supported_perimeter * 500

  return support_bonus - height_penalty + left_bonus
\end{lstlisting}

\section{Example generation prompt}
\label{app:prompt}

Listing~\ref{lst:prompt} shows one complete generation prompt in the form in which it is sent to the language model, for the class of Example 1 (moderate aspect ratio, medium width, high aspect-ratio variance, high width variance, 10 items). The prompt consists of a system message, which carries the fixed instruction and the description of the instance distribution (\ding{174} in Section~\ref{sec:search}), and a user message, which carries the program skeleton: the imports of the specification, the two best programs drawn from the island renamed \texttt{score\_placement\_v0} and \texttt{score\_placement\_v1} in increasing order of fitness, the docstring of the second replaced by ``Improved version of \texttt{score\_placement\_v0}'', and the header of \texttt{score\_placement\_v2}, whose body the model is asked to complete. For illustration, the two elites are the Best-Fit seed (v0) and the Lowest-Top seed (v1); during the search, they are whichever programs the cluster sampling draws. The model returns only the body of \texttt{score\_placement\_v2}. The FunSearch reference of Section~\ref{sec:benchmark} sends a similar prompt without the ``Problem context'' paragraph, and with a different signature of the scoring function.

\begin{lstlisting}[language={},caption={An exemplary generation prompt for the class of Example 1: system message and user message as sent to the model.},label={lst:prompt}]
=== system message ===
You are an expert Python programmer specialising in combinatorial optimisation.
Complete ONLY the body of the function shown. Output valid Python indented with 2 spaces.
Do NOT repeat the function header. Do NOT include markdown fences or any commentary.

Problem specification:
- Items arrive online (no lookahead, no relocation, 90-degree rotation allowed).
- Goal: minimise the final strip height.

Problem context (the instance distribution you are optimising for):  
- Width ~ Normal(mean=0.15, sd=0.075); 
- Aspect ratio h/w ~ Normal(mean=2.0, sd=1.0);
- Episode length: 10 items;
- Strip width: 1 unit (unbounded height). 

=== user message ===
import numpy as np
import math

def score_placement_v0(strip_width, placed_items, candidate, all_candidates):
  """Score a single candidate placement. Higher is better.

  Args:
    strip_width:    int  - width of the strip
    placed_items:   list of dicts with keys x, y, w, h
    candidate:      dict with keys x, y, w, h, rid - ONE candidate to score
    all_candidates: list of dicts - ALL candidates being considered

  Returns:
    float - higher score means better placement
  """
  current_max = max((p["y"] + p["h"] for p in placed_items), default=0)
  height_increase = max(0, candidate["y"] + candidate["h"] - current_max)
  return -(height_increase * 10000 + candidate["y"] * 100 + candidate["x"])


def score_placement_v1(strip_width, placed_items, candidate, all_candidates):
  """Improved version of `score_placement_v0`."""
  top = candidate["y"] + candidate["h"]
  min_top = min(c["y"] + c["h"] for c in all_candidates)
  max_top = max(c["y"] + c["h"] for c in all_candidates)
  span = max_top - min_top + 1
  return -(top - min_top) / span * 10000 - candidate["x"]


def score_placement_v2(strip_width, placed_items, candidate, all_candidates):
  """Improved version of `score_placement_v1`."""
\end{lstlisting}

\end{document}